\documentclass[journal]{IEEEtran}

\usepackage{amsmath,amsfonts}
\usepackage{algorithmic}
\usepackage{algorithm}
\usepackage{array}
\usepackage[caption=false,font=footnotesize]{subfig}
\usepackage{textcomp}
\usepackage{stfloats}
\usepackage{url}
\usepackage{verbatim}
\usepackage{graphicx}
\usepackage{subcaption}
\usepackage[backend=biber,style=ieee]{biblatex}
\DeclareBibliographyCategory{cited}
\AtEveryCitekey{\addtocategory{cited}{\thefield{entrykey}}}
\usepackage{graphicx}
\usepackage{comment}                              
\usepackage[super]{nth}

\usepackage[version=4]{mhchem}
\usepackage[utf8]{inputenc}
\usepackage{siunitx}
\usepackage{amsmath,amssymb,amsthm,mathtools}
\usepackage{longtable,tabularx}
\usepackage{bm}
\usepackage{commath}
\usepackage{thmtools}
\usepackage{mdframed}

\usepackage{longtable,tabularx}

\usepackage{comment}
\usepackage[super]{nth}
\newtheorem{definition}{Definition}[section]
\newtheorem{remark}{Remark}[section]

\newtheorem{assumption}{Assumption}[section]

\newtheorem{scenario}{Scenario}[section]

\usepackage{hyperref}

\usepackage{lipsum}

\hypersetup{
    colorlinks=false,
    urlbordercolor={1 0 0},
    pdfborder={0 0 1}
}

\newcommand{\R}{\mathbb{R}}

\newcommand{\sx}[1]{[#1]_{\times}}

\newcommand{\Tr}{\operatorname{tr}}

\ifCLASSINFOpdf
\else
\fi

\begin{document}

\title{TANGO-VIO: Triangulation-Aware Navigation with Guaranteed
Feature-Observability for Visual-Inertial Odometry}

\author{%
Abdulbaki Sanlan,
Ege C. Altunkaya%
\thanks{Abdulbaki Sanlan and Ege C. Altunkaya contributed equally to this work.},
Hasan Taha Bagci,
Emre Koyuncu,~\IEEEmembership{Senior Member,~IEEE},
and~Ibrahim Ozkol%
\thanks{Abdulbaki Sanlan, Emre Koyuncu, and Ibrahim Ozkol are with the
Faculty of Aeronautics and Astronautics, Istanbul Technical University,
34467 İstanbul, Türkiye.}%
\thanks{Ege C. Altunkaya is with the Aviation Institute, Istanbul Technical
University, 34467 İstanbul, Türkiye.}%
\thanks{Hasan T. Bagci is with the Aerospace Research Center, Istanbul
Technical University, 34467 İstanbul, Türkiye.}%
}

\maketitle

\begin{abstract}
    In vision-aided navigation and visual-inertial odometry, the quality of
    triangulated three-dimensional feature positions is a fundamental prerequisite for state estimation accuracy. Triangulation becomes ill-conditioned or even impossible when a camera undergoes pure rotation without translation, or when the observed bearing vectors provide insufficient parallax. Even though visual-inertial odometry has been extensively studied, the active maintenance of feature-observability during navigation has not been sufficiently addressed in the literature. To address this gap, this study presents \textbf{TANGO-VIO}, a triangulation-aware navigation framework that embeds a log-determinant metric of the feature-wise stacked-bearing matrix into a control barrier function. In this proposed method, the observability guarantee is established in the feature-geometric sense by enforcing a lower bound on the aggregate triangulation-information metric through a nominal-direction-weighted minimum-deviation velocity correction. The proposed architecture is evaluated through software-in-the-loop simulations and real flight experiments. The results show improved triangulation conditioning under low-parallax motion, while the flight response closely reproduces the corresponding simulation behavior and confirms the practical realizability of the proposed safety filter. Supplementary materials are available on the \href{https://tango-vio.github.io/}{project webpage}.
\end{abstract}


\IEEEpeerreviewmaketitle

\section{Introduction}
\label{introduction}

\IEEEPARstart{V}{isual-inertial} odometry (VIO) has emerged as a fundamental enabling method for autonomous navigation in GPS-denied environments. By fusing measurements from a monocular or stereo camera with data from an inertial measurement unit (IMU), VIO systems provide real-time estimates of six-degree-of-freedom pose, velocity, and sensor biases for a broad class of robotic platforms, including aerial vehicles, ground robots, and consumer augmented-reality devices. The complementary nature of visual and inertial sensing---the camera observes scene structure to constrain drift, while the IMU propagates the state between image frames at high rate---underpins the success of both Kalman Filter-based architectures \cite{mourikis2007msckf, strasdat2010real, li2013high, bloesch2017iterated, csanlan2025particle, wang2025po} and nonlinear-optimization-based pipelines \cite{leutenegger2015okvis, qin2018vinsmono, rosinol2020kimera, campos2021orb, luo2025supervins}. Each paradigm presents distinct advantages and disadvantages in terms of accuracy, computational cost, and robustness \cite{delmerico2018benchmark}.

Despite its maturity, VIO remains acutely sensitive to the quality of 3D feature triangulation and to the observability of its internal states. Accurate triangulation requires sufficient baseline (parallax) between the camera poses from which a landmark is observed; when the platform undergoes pure rotation, near-static operation, or translational motion that produces insufficient angular separation between successive bearing rays, the stacked-bearing matrix becomes ill-conditioned or rank deficient \cite{wu2017unobs}. These motions can also enlarge the unobservable subspace of the visual-inertial system beyond its inherent four-dimensional nullspace---global translation and yaw---introducing additional ambiguities in metric scale and attitude \cite{hesch2014consistency, wu2017unobs, hernandez2015observability}. Hailu and Gebregziabher \cite{hailuquantifying} quantified this trajectory dependence for metric scale, showing that curved trajectories yield significantly better scale observability than straight-line motion. These results collectively underscore that the state observability and feature-observability of VIO are not fixed system properties but depend critically on the trajectory executed by the platform.

A natural countermeasure to observability loss is to plan trajectories that explicitly maximize estimation quality. Hausman et al.\ \cite{hausman2016obsaware} introduced an observability-aware trajectory-optimization framework that maximizes the smallest singular value of a local observability Gramian to ensure that states such as IMU biases remain estimable. Grebe et al.\ \cite{grebe2021obsaware} extended this line of work with a comprehensive comparison of deterministic and stochastic observability metrics for extrinsic self-calibration, concluding that both families significantly outperform random and minimum-energy trajectories. While powerful, these methods operate in an open-loop, offline fashion: they pre-compute an entire trajectory 
before execution and cannot be recomputed at the control rate, making them unsuitable for dynamic environments or real-time reference tracking. More closely related to online operation, Wu et al.\ \cite{wu2022perception} proposed a receding-horizon perception-aware planner for multicopters that samples minimum-jerk trajectories and selects the one minimizing a perception cost derived from the predicted pose-estimation covariance. Their formulation is notable for incorporating
motion-blur effects on visual features; nevertheless, the sampling-based evaluation limits the update rate and provides no formal guarantee that the selected trajectory maintains a minimum level of feature-geometry conditioning. From a control-theoretic perspective, Ogri et al.\ \cite{ogri2024adaptive} formulated an adaptive optimal controller that simultaneously drives a monocular camera toward a goal position and maximizes depth observability by incorporating an orthogonality-based information term into the cost function, and later extended the framework to constrained spacecraft inspection \cite{ogri2025monocular}. However, these control-based formulations assume planar feature geometry, do not extend to the full three-dimensional triangulation problem encountered in VIO, and encode the observability-related objective as a soft cost rather than a hard safety constraint.

Several complementary strategies have been explored to mitigate specific degeneracy modes. On the estimator side, zero-velocity updates bound drift during stationary intervals by constraining the velocity state to zero \cite{ramanandan2011inertial}, while Kottas et al.\ \cite{kottas2013detecting} detect hovering and freeze the sliding-window clones to preserve parallax until translational motion resumes. A common hardware-based alternative is to augment the pipeline with a range sensor: Delaune et al.\ \cite{delaune2021rangevio} showed that a one-dimensional laser range finder eliminates metric-scale ambiguity even under constant-velocity flight, assuming locally flat terrain; Alberico et al.\ \cite{alberico2024structure} later relaxed this assumption with a structure-invariant formulation valid for arbitrary terrain profiles. Xu et al.\ \cite{xu2023novel} recovered scale from a single-point laser via calibrated camera--laser geometry and background tracking, and Huo and Liu \cite{huo2023range} incorporated range factors into a factor-graph VIO initialization to handle insufficiently excited motion. In the underwater domain, Ding et al.\ \cite{ding2023rd} fused a depth gauge with VIO as a unary vertical constraint. While these range- and depth-aided approaches restore observability without demanding specific motion excitation, they introduce additional hardware cost, weight, and calibration burden. 

In parallel, Control Barrier Functions (CBFs) have gained prominence as a principled framework for enforcing state constraints in safety-critical robotic systems by rendering a user-defined safe set forward invariant \cite{ames}. In visual robotics, CBFs have been applied predominantly to enforce visibility constraints---ensuring that target features remain within the camera field of view. Salehi et al.\ \cite{salehi2021cbfibvs} developed a barrier-function-based transformation that maps image-plane feature dynamics into an equivalent unconstrained system, guaranteeing that all feature points satisfy user-prescribed bounds throughout the servoing task. Heshmati-Alamdari et al.\ \cite{heshmati2024cbfmms} proposed a two-tiered CBF-based image-based visual servoing (IBVS) controller for mobile manipulator systems that simultaneously enforces field-of-view constraints and joint-level velocity limits. Abdi et al.\ \cite{abdi2023vcbf} introduced a vision-based CBF (V-CBF) constructed directly from RGB-D images using a conditional generative adversarial network, enabling safe navigation in unknown environments with obstacles of arbitrary shape. In the domain of observability enforcement, Coleman et al.\ \cite{coleman2024cbfobs} demonstrated that a CBF can maintain a lower bound on the inverse condition number of the Fisher information matrix for a range-based target-tracking problem, preventing the tracker from entering geometrically degenerate configurations. While these works establish the utility of CBFs for visibility and observability constraints, none addresses the specific problem of maintaining triangulation quality, thereby feature-observability, in a visual-inertial navigation context. The interplay between the camera's bearing measurements, the platform's translational motion, and the resulting information content of the triangulation subproblem has not been encoded as a CBF constraint.

\subsection{Problem Statement}
\label{problemStatement}

The foregoing survey reveals a significant gap. Observability-aware trajectory-optimization methods \cite{hausman2016obsaware, grebe2021obsaware} generate motions that improve state estimation quality but operate offline. Perception-aware planners \cite{wu2022perception} and adaptive optimal controllers \cite{ogri2024adaptive, ogri2025monocular} operate online but encode observability-related objectives as soft costs without providing a hard guarantee on minimum triangulation quality. Range- and depth-aided approaches \cite{delaune2021rangevio, alberico2024structure, xu2023novel, huo2023range, ding2023rd} restore metric-scale observability but at the expense of additional sensing hardware and without directly regulating triangulation conditioning. CBF-based methods have been successfully applied to field-of-view constraints \cite{salehi2021cbfibvs, heshmati2024cbfmms} and to observability of range-based tracking \cite{coleman2024cbfobs}, but have not yet been brought to bear on the triangulation problem central to visual-inertial odometry.

Building on this gap, a pertinent research question arises: \emph{Can feature-observability be enforced as a real-time safety constraint while retaining the nominal trajectory as a minimum-deviation reference, rather than replacing it with an offline observability-optimized trajectory?} In this context, feature-observability refers to the feature-geometric conditioning of the bearing-based triangulation problem, rather than to a complete nonlinear observability analysis of the full VIO state.

The primary objective of this study is to develop a triangulation-aware
navigation layer that actively maintains sufficient parallax during
VIO-guided motion while preserving the nominal velocity command whenever
it is compatible with the triangulation-quality requirement. 
To achieve this objective, \textbf{TANGO-VIO} (\textbf{T}riangulation-\textbf{A}ware \textbf{N}avigation with \textbf{G}uaranteed Feature-\textbf{O}bservability for \textbf{V}isual-\textbf{I}nertial \textbf{O}dometry) 
is introduced as a navigation-level safety-filtering framework. 
Here, the guaranteed feature-observability component refers to 
CBF-certified preservation of a lower-bounded feature-geometric
triangulation-information metric. The framework interfaces with 
any VIO pipeline that maintains a sliding window of camera poses
with multi-view feature tracks, and incorporates the following
key components:
\begin{enumerate}
    \item \emph{Log-determinant CBF construction}: to encode the aggregate triangulation quality as a safety constraint with a prescribed lower bound;
    \item \emph{Hard and soft CBF operation}: to represent both strict triangulation-priority behavior and mission-priority behavior through a slack-enabled relaxation mechanism.
\end{enumerate}

The use of CBFs is motivated by their ability to enforce state-dependent constraints online through computationally efficient optimization. In contrast to offline observability-aware trajectory generation, the proposed formulation acts directly at the velocity-command layer and modifies the nominal command only when the current bearing geometry requires additional parallax. 

The proposed methodology is validated through software-in-the-loop simulations and real flight experiments, focusing on triangulation-quality preservation, velocity-command modification, trajectory-level deviation, and the resulting visual-information quality.

\subsection{Contributions and Organization}
\label{contributionsOrganization}

The contributions of this study are itemized as follows:

\begin{itemize}

\item Triangulation-aware navigation is formulated as an online velocity-filtering problem for visual-inertial odometry. Instead of treating feature-observability as a passive byproduct of the executed motion, the proposed framework actively modifies the commanded motion when the parallax becomes insufficient for well-conditioned triangulation.

\item Aggregate feature triangulation quality is encoded as a Control Barrier Function constraint. In particular, the average log-determinant of the feature-wise stacked-bearing matrices is used to define a safe set that lower-bounds the aggregate triangulation informativeness over the active camera-pose window.

\item A minimum-deviation safety filter is formulated with directional weighting that penalizes deviations perpendicular to the nominal velocity more strongly than deviations along it. The formulation supports both a hard-CBF mode, which strictly prioritizes triangulation-quality preservation, and a slack-enabled soft-CBF mode, which permits controlled constraint relaxation to better preserve the incoming nominal velocity command.

\end{itemize}

The remainder of the paper is organized as follows: Section~\ref{preliminaries} establishes the mathematical preliminaries on bearing-based triangulation and control barrier functions. Section~\ref{methodology} derives the triangulation-aware CBF and the associated QP formulation. Section~\ref{results} presents simulation and flight-test results. Finally, Section~\ref{conclusion} offers concluding remarks and future directions.

\section{Preliminaries}
\label{preliminaries}

The necessary background for the subsequent sections is established in this section.

\subsection{Notations}
\label{notations}

A vector is denoted by bold lowercase type, e.g., $\bm{x}$, while a matrix is denoted by bold uppercase type, e.g., $\bm{A}$. The transpose of a vector or matrix is denoted by $(\cdot)^{\top}$, and the time derivative of a variable is denoted by $\dot{(\cdot)}$. The Euclidean norm is denoted by $\|\cdot\|_{2}$. For a symmetric positive-definite matrix $\bm{W}$, the weighted norm is denoted by $\|\cdot\|_{\bm{W}}$, where $\|\bm{a}\|_{\bm{W}}^{2} \triangleq \bm{a}^{\top}\bm{W}\bm{a}$. The trace and determinant operators are denoted by $\Tr(\cdot)$ and $\det(\cdot)$, respectively. The set of real numbers is denoted by $\mathbb{R}$, and the identity matrix of compatible dimension is denoted by $\bm{I}$.

Coordinate frames are denoted by calligraphic letters, e.g., $\mathcal{F}_{A}$. A vector $\bm{p}$ expressed in frame $\mathcal{F}_{A}$ is written as ${}^{A}\bm{p}$, where the superscript is the identifying label of the frame. When the point or object associated with the vector must be specified, it appears as a subscript, e.g., ${}^{A}\bm{p}_{C}$ denotes the position of point $C$ expressed in $\mathcal{F}_{A}$. A rotation matrix from $\mathcal{F}_{X}$ to $\mathcal{F}_{Y}$ is denoted by $\bm{R}_{X}^{Y}$, so that ${}^{Y}\bm{a}=\bm{R}_{X}^{Y}{}^{X}\bm{a}$. The skew-symmetric matrix associated with $\bm{a}\in\mathbb{R}^{3}$ is denoted by $\sx{\bm{a}}$ and satisfies $\sx{\bm{a}}\bm{b}=\bm{a}\times\bm{b}$. 

The following frames are used throughout this paper. $\mathcal{F}_{N}$ is the navigation frame, a local-level Earth-fixed frame with East--North--Up (ENU) orientation. $\mathcal{F}_{B}$ is the body frame, rigidly attached to the vehicle with Forward--Left--Up (FLU) convention. $\mathcal{F}_{C}$ is the camera frame, with the $z$-axis along the principal ray; camera poses in the active sliding window are indexed by $j=1,\ldots,n$, where $C_j$ denotes the $j$-th historical pose and $C_n$ is the most recent. $\mathcal{F}_{A}$ is the anchor frame, which coincides with $C_1$ and is held fixed over the window; all bearing vectors and feature positions are expressed in $\mathcal{F}_{A}$. Visual features are indexed by $i=1,\ldots,s$, where $s$ and $n$ denote the total number of tracked features and camera poses in the active window, respectively. A finite indexed collection of elements is written as $\{\cdot\}_{k=1}^{m}$, where $k$ is the running index and $m$ is the cardinality.

For a scalar function $h(\bm{x})$, the gradient with respect to $\bm{x}$ is denoted by $\nabla_{\bm{x}}h(\bm{x})$. The Lie derivative of $h$ along a vector field $\bm{f}(\bm{x})$ is denoted by

\begin{equation}
    \mathcal{L}_{\bm{f}}h(\bm{x})
    \triangleq
    \nabla_{\bm{x}}h(\bm{x})^{\top}\bm{f}(\bm{x}).
\end{equation}
Finally, a relative-degree-one control-affine system is written as
\begin{equation}
\label{controlAffineSystem}
    \dot{\bm{x}} = \bm{f}(\bm{x}) + \bm{g}(\bm{x})\bm{u},
\end{equation}
where $\bm{x} \in \mathbb{R}^n$ is state vector, $\bm{u} \in \mathbb{R}^m$ is control input vector. Nonlinear mappings of $\bm{f} : \mathbb{R}^n \to \mathbb{R}^n$ and $\bm{g} : \mathbb{R}^n \to \mathbb{R}^{n \times m}$ are locally Lipschitz continuous functions.

\subsection{Control Barrier Functions}
\label{controlBarrierFunctions}

To encode state constraints, let $h : D\subset\mathbb{R}^n \to \mathbb{R}$ be a continuously differentiable barrier function defined on a domain $D$. The associated safe set is

\begin{equation}
    C \triangleq \{\bm{x}\in D \mid h(\bm{x}) \ge 0\},
\end{equation}
with boundary $\partial C \triangleq \{\bm{x}\in D \mid h(\bm{x})=0\}$ and interior $\mathrm{Int}(C)\triangleq \{\bm{x}\in D \mid h(\bm{x})>0\}$. The control objective is to render $C$ forward invariant. Consider the control-affine system in~\eqref{controlAffineSystem} with admissible inputs $\bm{u}\in U\subset\mathbb{R}^m$. The Lie derivatives of $h$ along $\bm f$ and $\bm g$ are

\begin{equation}
\begin{split}
    \mathcal{L}_f h(\bm{x}) \triangleq \nabla h(\bm{x})^\top \bm{f}(\bm{x}),& \\
    \mathcal{L}_g h(\bm{x}) \triangleq \nabla h(\bm{x})^\top \bm{g}(\bm{x}).&
\end{split}
\end{equation}

\begin{definition} [Control Barrier Function \cite{ames}]
    A continuously differentiable function $h:D\to\mathbb{R}$ is a control barrier function on $D$ if there exists an extended class-$\mathcal{K}_\infty$ function $\alpha$ such that, for all $\bm{x} \in D$,
    
    \begin{equation}
    \label{cbf1}
        \sup_{\bm{u}\in U}\Big\{\mathcal{L}_f h(\bm{x})+\mathcal{L}_g h(\bm{x})\,\bm{u}\Big\}
        \ge -\alpha\!\big(h(\bm{x})\big).
    \end{equation}
    
    Equivalently, the set of CBF-admissible controls
    \begin{equation}
    \label{cbf2}
        \mathcal{U}_h(\bm{x}) \triangleq \Big\{\bm{u}\in U \,\big|\, \mathcal{L}_f h(\bm{x})+\mathcal{L}_g h(\bm{x})\,\bm{u}+\alpha\!\big(h(\bm{x})\big)\ge 0\Big\}
    \end{equation}
    
is nonempty for all $\bm{x}\in D$.

\end{definition}

\subsection{Multi-View Feature Tracks in Sliding-Window Visual-Inertial Odometry}
\label{slidingWindowFeatureTracks}

VIO pipelines, whether filter-based~\cite{mourikis2007msckf,li2013optimization,delaune2020xvio,geneva2020openvins} or optimization-based over a factor graph~\cite{leutenegger2015okvis,qin2018vinsmono}, share a common structural element: a sliding window of $n$ historical camera poses $\{C_1,\ldots,C_n\}$ together with visual features tracked across multiple poses within this window. The TANGO-VIO framework is built exclusively on this common structure and is therefore agnostic to the underlying estimator.

The sole structural requirement of the proposed formulation is the availability of \emph{multi-view feature tracks}: features observed from at least two camera poses in the active window. Any such feature is a candidate for (or subject to) bearing-based triangulation and contributes a set of anchor-frame bearing vectors from which its stacked-bearing matrix is constructed. The barrier function proposed in this work can therefore be evaluated on any subset of $s$ tracked features satisfying this requirement, independent of how the estimator classifies, parameterizes, or marginalizes those features internally.

\subsubsection*{Active Camera-Pose Window and Bearing Accumulation}
The active window contains the $n$ most recent pose states $\{{}^{A}\bm{p}_{C_j},\,\bm{R}_{C_j}^{A}\}_{j=1}^{n}$. For each tracked feature $i$, the set of anchor-frame bearing vectors $\{\bm{b}_{i,j}\}_{j=1}^{n}$ accumulated over this window encodes the multi-view parallax geometry that determines how well-conditioned the feature's triangulation is. TANGO-VIO monitors and actively shapes this geometry by treating the aggregate feature-observability of the multi-view tracked set as a real-time safety constraint, preventing the VIO front-end from entering degenerate low-parallax regimes.

\subsection{Stacked-Bearing Matrix}
\label{triangulationInformationMatrix}

\begin{figure}[!t]
    \centering
    \includegraphics[width=\columnwidth]{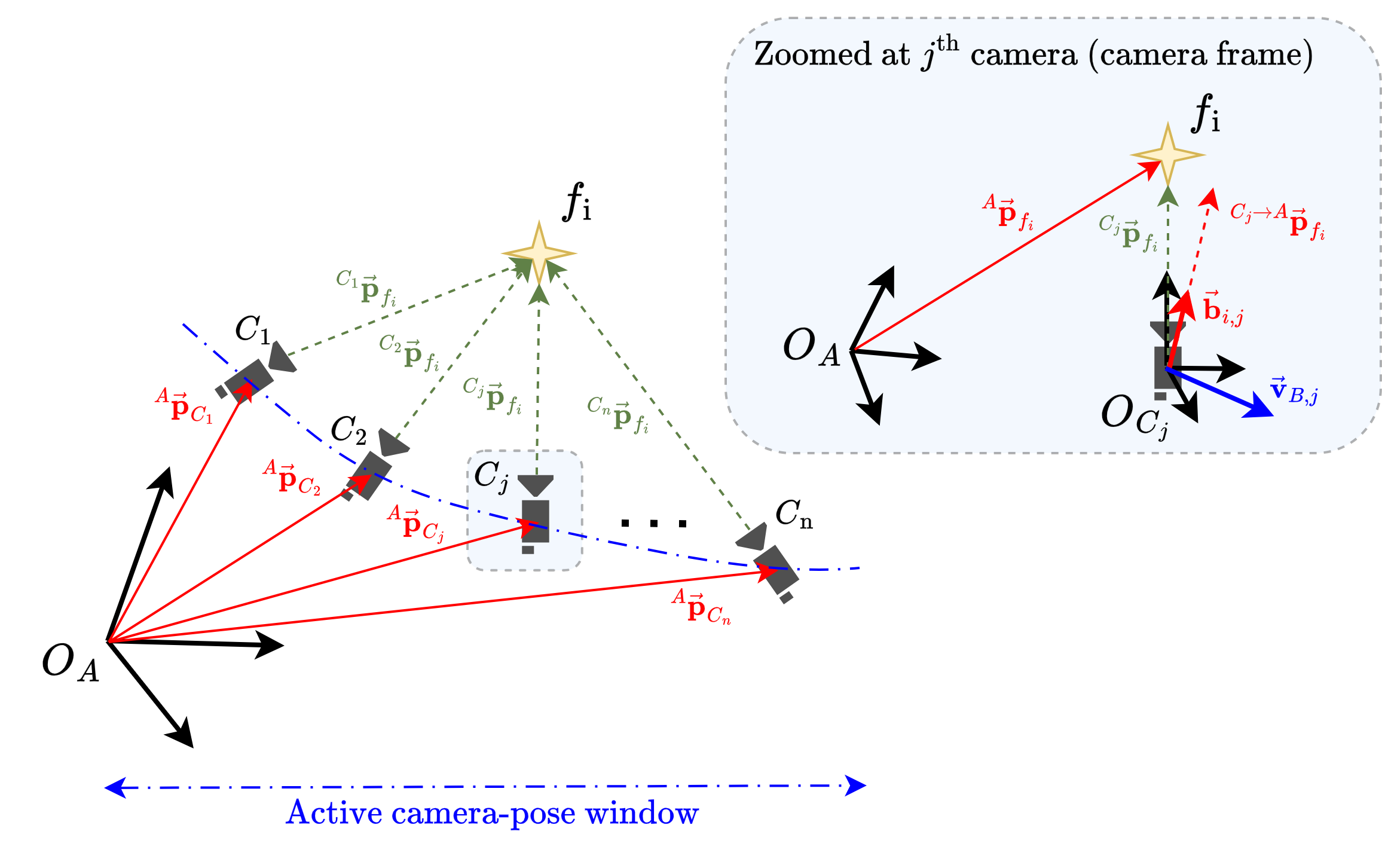}
    \caption{Bearing-vector geometry over the active camera-pose window. A stationary feature $f_i$ is observed from multiple camera poses $C_j$. For each observation, the anchor-frame relative vector ${}^{C_j \to A}\bm{p}_{f_i}={}^{A}\bm{p}_{f_i}-{}^{A}\bm{p}_{C_j}$ lies along the corresponding bearing ray $\bm{b}_{i,j}$. The body and camera frames have coincident origins, while a possible rotation between them is represented by $\bm{R}_{B}^{C}$. Accordingly, the body-frame velocity $\bm{v}_{B,j}$ is mapped into the anchor frame through $\bm{R}_{C_j}^{A}\bm{R}_{B}^{C}$ to characterize its effect on the bearing geometry. This relation is subsequently used to derive the bearing-rate dynamics required to differentiate the stacked-bearing matrix.}
    \label{fig:bearingVector}
\end{figure}

Consider a stationary feature $f_i$ observed from the active camera-pose window
$\{C_j\}_{j=1}^{n}$. The corresponding bearing geometry is illustrated in
Fig.~\ref{fig:bearingVector}. For the observation obtained at camera pose
$C_j$, the feature position expressed in the camera frame is
\begin{equation}
  {}^{C_j}\bm{p}_{f_i}
  =
  \bm{R}_{A}^{C_j}
  \left(
  {}^{A}\bm{p}_{f_i}
  -
  {}^{A}\bm{p}_{C_j}
  \right).
  \label{eq:feature_camera_frame}
\end{equation}
where ${}^{C_j}\bm{p}_{f_i}\in\mathbb{R}^{3}$ denotes the feature position
expressed in $\mathcal{F}_{C_j}$, whereas ${}^{A}\bm{p}_{f_i}$ and
${}^{A}\bm{p}_{C_j}$ are expressed in the anchor frame $\mathcal{F}_{A}$. Rotating \eqref{eq:feature_camera_frame} back to the anchor frame gives the
relative feature--camera vector
\begin{equation}
  \begin{split}
  {}^{C_j\to A}\bm{p}_{f_i}
  &=
  \bm{R}_{C_j}^{A}\,{}^{C_j}\bm{p}_{f_i} \\
  &=
  {}^{A}\bm{p}_{f_i}
  -
  {}^{A}\bm{p}_{C_j} \\
  &=
  \rho_{i,j}\bm{b}_{i,j}.
  \end{split}
  \label{eq:feature_anchor_factored}
\end{equation}
The vector $\bm{b}_{i,j}\in\mathbb{R}^{3}$ denotes the bearing of feature
$f_i$ observed from pose $C_j$ and expressed in $\mathcal{F}_{A}$, and
$\rho_{i,j} \in \mathbb{R}_{>0}$ is the corresponding range scale. For non-unit bearing representations, $\bm{b}_{i,j}=\eta_{i,j}\hat{\bm{b}}_{i,j}$ with $\eta_{i,j}=\|\bm{b}_{i,j}\|_2$ and $\|\hat{\bm{b}}_{i,j}\|_2=1$; for
unit-normalized bearings, $\eta_{i,j}=1$.

Since the feature, camera center, and bearing ray are collinear, the
cross-product constraint for pose $C_j$ is
\begin{equation}
  \sx{\bm{b}_{i,j}}
  \left(
  {}^{A}\bm{p}_{f_i}
  -
  {}^{A}\bm{p}_{C_j}
  \right)
  =
  \bm{0}.
  \label{eq:single_bearing_cross_constraint}
\end{equation}
With $\bm{N}_{i,j}\triangleq\sx{\bm{b}_{i,j}}$, stacking the constraints from
all $n$ camera poses yields the linear least-squares triangulation system
\begin{equation}
  \underbrace{
  \begin{bmatrix}
  \bm{N}_{i,1}\\
  \vdots\\
  \bm{N}_{i,n}
  \end{bmatrix}}_{\bm{A}_i}
  {}^{A}\bm{p}_{f_i}
  =
  \underbrace{
  \begin{bmatrix}
  \bm{N}_{i,1}{}^{A}\bm{p}_{C_1}\\
  \vdots\\
  \bm{N}_{i,n}{}^{A}\bm{p}_{C_n}
  \end{bmatrix}}_{\bm{q}_i}.
  \label{eq:stacked_bearing_system}
\end{equation}
where $\bm{A}_{i}\in\mathbb{R}^{3n\times3}$ and
$\bm{q}_{i}\in\mathbb{R}^{3n}$. The least-squares estimate satisfies
\begin{equation}
  \bm{A}_{i}^{\top}\bm{A}_{i}{}^{A}\bm{p}_{f_i}
  =
  \bm{A}_{i}^{\top}\bm{q}_{i}.
  \label{eq:stacked_bearing_normal_equation}
\end{equation}

The normal matrix $\bm{A}_{i}^{\top}\bm{A}_{i}$ depends only on the bearing
geometry accumulated over the active window. Using the skew-symmetric identity,
\begin{equation}
  \begin{split}
  \bm{N}_{i,j}^{\top}\bm{N}_{i,j}
  &=
  \sx{\bm{b}_{i,j}}^{\top}\sx{\bm{b}_{i,j}} \\
  &=
  \eta_{i,j}^{2}\bm{I}
  -
  \bm{b}_{i,j}\bm{b}_{i,j}^{\top} \\
  &\triangleq
  \bm{\pi}_{i,j}.
  \end{split}
  \label{eq:projection_matrix_def}
\end{equation}
The matrix $\bm{\pi}_{i,j}$ is the bearing-induced projection matrix; it
removes the component along the bearing ray and retains the transverse
information contributed by the observation.

The per-feature stacked-bearing matrix is therefore defined as
\begin{equation}
  \begin{split}
  \bm{M}_{i}
  &\triangleq
  \bm{A}_{i}^{\top}\bm{A}_{i} \\
  &=
  \sum_{j=1}^{n}\bm{\pi}_{i,j} =
  \sum_{j=1}^{n}
  \left(
  \eta_{i,j}^{2}\bm{I}
  -
  \bm{b}_{i,j}\bm{b}_{i,j}^{\top}
  \right).
  \end{split}
  \label{eq:Mi_def}
\end{equation}
where $\bm{M}_{i}\in\mathbb{R}^{3\times3}$. Thus, $\bm{M}_{i}$ summarizes the
multi-view bearing geometry available for triangulating feature $f_i$.
Well-distributed bearing rays yield a well-conditioned matrix, whereas nearly
parallel bearing rays lead to poor conditioning or rank deficiency. Therefore, $\bm{M}_{i}$ is used as the feature-wise stacked-bearing matrix.

\begin{remark}
The matrix $\bm{M}_{i}$ is positive semidefinite because it is the normal
matrix of the stacked least-squares system. It becomes positive definite only
when the bearing set $\{\bm{b}_{i,j}\}_{j=1}^{n}$ provides sufficient parallax to constrain the three-dimensional feature position. Hence, each tracked feature has its own stacked-bearing matrix and its own
rank-deficiency risk.
\end{remark}

\section{Methodology}
\label{methodology}

\begin{figure*}[!t]
    \centering
    \includegraphics[width=\textwidth]{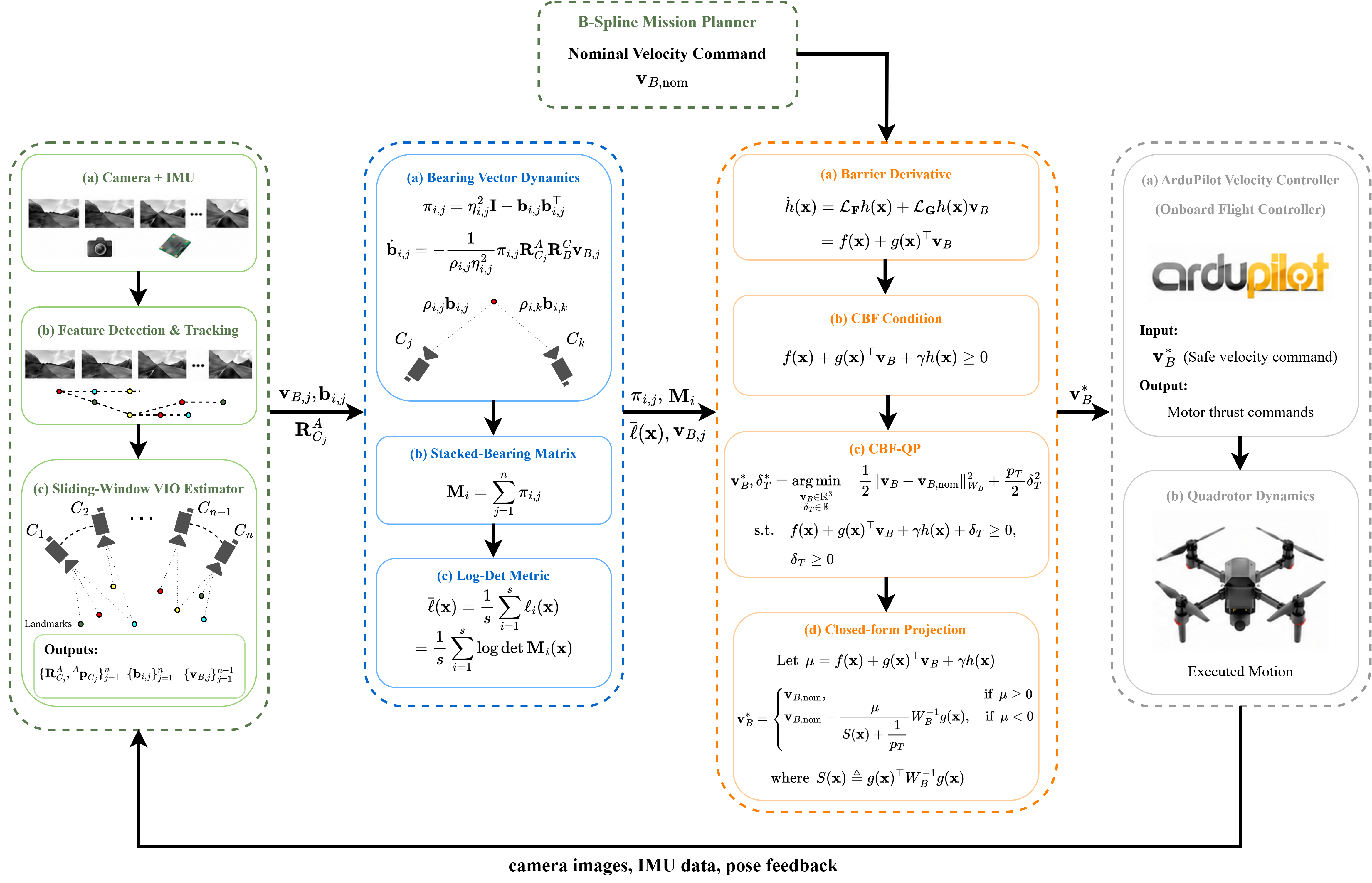}
    \caption{Information flow of the proposed TANGO-VIO architecture: Camera images and IMU measurements are processed by the VIO front end to detect and track visual features, maintain the active sliding window of camera poses, and provide the bearing vectors $\bm{b}_{i,j}$, pose transformations, and recorded body-frame velocities associated with the window. These quantities are passed to the triangulation-geometry module, where the bearing-rate dynamics are evaluated, the projection matrices $\bm{\pi}_{i,j}$ are constructed, and the feature-wise stacked-bearing matrices $\bm{M}_{i}$ are obtained. The resulting matrices are aggregated through the average log-det metric $\bar{\ell}(\bm{x})$, which defines the triangulation-quality barrier. In parallel, the B-spline mission planner supplies the nominal body-frame velocity command $\bm{v}_{B,\mathrm{nom}}$. The barrier derivative module combines the triangulation metric with the bearing dynamics to obtain the control-affine form $\dot{h}(\bm{x})=f(\bm{x})+\bm{g}(\bm{x})^{\top}\bm{v}_{B}$, which is injected into the CBF condition and the associated CBF-QP. The safety filter then projects the nominal velocity command onto the triangulation-admissible set, or relaxes the constraint through the slack variable in soft-CBF operation, producing the corrected command $\bm{v}_{B}^{*}$. This command is sent to the ArduPilot velocity controller, which generates motor thrust commands for the quadrotor. The executed motion produces new camera, IMU, and pose feedback, closing the loop for the next safety-filter update.}
    \label{fig:generalOverview}
\end{figure*}

The proposed TANGO-VIO methodology is summarized in Fig.~\ref{fig:generalOverview}. The framework operates as a navigation-level safety filter around a standard VIO-guided drone architecture. Onboard camera and IMU measurements are first processed by the VIO front end to obtain tracked features, sliding-window camera poses, and the visual-geometry quantities required for evaluating the triangulation-quality barrier. In parallel, an upstream command source provides the nominal body-frame velocity command $\bm{v}_{B,\mathrm{nom}}$. In the present experiments, this command is generated by a B-spline mission planner to ensure repeatable reference motions; however, the proposed formulation is not tied to this specific planner. The same command-filtering structure can also be used with a higher-level autonomy module or with human-pilot velocity commands supplied through a remote-control interface. The active camera-pose window is used to construct the feature-wise stacked-bearing matrices, from which the average log-determinant metric and the corresponding CBF candidate are evaluated. The TANGO-VIO safety filter then enforces this metric through a CBF-QP, preserving the incoming nominal command whenever it satisfies the triangulation-quality constraint and otherwise applying the minimum weighted correction required to generate additional parallax. The resulting safe velocity command is finally sent to the ArduPilot velocity controller for drone execution.

\subsection{Bearing Vector Dynamics}
\label{bearingVectorDynamics}

The bearing dynamics provide the connection between platform motion and the
evolution of triangulation geometry. In the proposed formulation, the
stacked-bearing matrix is constructed from the bearing-dependent
projection matrices associated with the active camera-pose window. Therefore,
to obtain a control-affine derivative of the triangulation-quality metric, it is necessary to express how the bearing vector changes as a function of the body-frame translational velocity command. The following derivation specializes this bearing-rate relationship to the anchor-frame bearing convention and the body-frame velocity input used in the TANGO-VIO formulation.

The bearing dynamics are derived from the relative feature--camera geometry
defined in \eqref{eq:feature_camera_frame}--\eqref{eq:feature_anchor_factored}. For compactness, define the anchor-frame relative position vector from camera pose $C_j$ to feature $f_i$ as
\begin{equation}
    \begin{split}
  {}^{A}\bm{r}_{i,j}
  &\triangleq
  {}^{C_j\to A}\bm{p}_{f_i}, \\
  &=
  {}^{A}\bm{p}_{f_i}-{}^{A}\bm{p}_{C_j}, \\
  &= \rho_{i,j}\bm{b}_{i,j}.
  \label{eq:relative_position_anchor}
\end{split}
\end{equation}
The pose-dependent translational kinematics of the camera are written as
\begin{equation}
  {}^{A}\dot{\bm{p}}_{C_j}
  =
  \bm{R}_{C_j}^{A}\bm{R}_{B}^{C}\bm{v}_{B,j},
  \label{eq:pose_velocity}
\end{equation}
where $\bm{v}_{B,j}$ denotes the body-frame translational velocity associated
with pose $j$.

\begin{assumption}[Stationary feature over the active window]
\label{assumption:static_feature}
The tracked point feature is assumed to be fixed in the anchor frame over the
active camera-pose window, i.e.,
\begin{equation}
  {}^{A}\dot{\bm{p}}_{f_i}=\bm{0}.
\end{equation}
This assumption is consistent with standard VIO operation, where tracked
landmarks are treated as static scene points over a short sliding window, while
independently moving features are rejected by the front-end tracking and
outlier-rejection stages.
\end{assumption}

Differentiating \eqref{eq:relative_position_anchor} with respect to time gives
\begin{align}
  {}^{A}\dot{\bm{r}}_{i,j}
  &=
  {}^{A}\dot{\bm{p}}_{f_i}
  -
  {}^{A}\dot{\bm{p}}_{C_j} \nonumber\\
  &=
  \dot{\rho}_{i,j}\bm{b}_{i,j}
  +
  \rho_{i,j}\dot{\bm{b}}_{i,j}.
  \label{eq:relative_position_derivative}
\end{align}
Using Assumption~\ref{assumption:static_feature}, this reduces to
\begin{equation}
  -{}^{A}\dot{\bm{p}}_{C_j}
  =
  \dot{\rho}_{i,j}\bm{b}_{i,j}
  +
  \rho_{i,j}\dot{\bm{b}}_{i,j}.
  \label{eq:relative_derivative_static_feature}
\end{equation}
Equation~\eqref{eq:relative_derivative_static_feature} separates the effect of
camera translation into two components: a radial component that changes the
depth $\rho_{i,j}$ and a transverse component that changes the bearing
direction. Since feature-observability depends on changes in bearing geometry,
the transverse component is the quantity that ultimately enters the derivative
of the information matrix.

The bearing vector is written as
$\bm{b}_{i,j}=\eta_{i,j}\hat{\bm{b}}_{i,j}$, where
$\|\hat{\bm{b}}_{i,j}\|_2=1$. The normalization scale $\eta_{i,j}$ is treated
as constant for each bearing observation; in particular, for unit-normalized
bearings, $\eta_{i,j}=1$. Therefore, multiplying \eqref{eq:relative_derivative_static_feature} by
$\hat{\bm{b}}_{i,j}^{\top}/\eta_{i,j}$ and using $\hat{\bm{b}}_{i,j}^{\top}\bm{b}_{i,j} = \eta_{i,j}$ and $\hat{\bm{b}}_{i,j}^{\top}\dot{\bm{b}}_{i,j} = 0$ yields
\begin{align}
  -\frac{\hat{\bm{b}}_{i,j}^{\top}}{\eta_{i,j}}
  {}^{A}\dot{\bm{p}}_{C_j}
  &=
  \dot{\rho}_{i,j}
  \frac{\hat{\bm{b}}_{i,j}^{\top}\bm{b}_{i,j}}{\eta_{i,j}}
  +
  \rho_{i,j}
  \frac{\hat{\bm{b}}_{i,j}^{\top}\dot{\bm{b}}_{i,j}}{\eta_{i,j}},
  \nonumber\\
  &=
  \dot{\rho}_{i,j}.
\end{align}
Thus, the depth-rate expression becomes
\begin{equation}
  \dot{\rho}_{i,j}
  =
  -\frac{\hat{\bm{b}}_{i,j}^{\top}}{\eta_{i,j}}
  {}^{A}\dot{\bm{p}}_{C_j}.
  \label{eq:rhodot_anchor_velocity}
\end{equation}
Substituting \eqref{eq:pose_velocity} into \eqref{eq:rhodot_anchor_velocity}
gives
\begin{equation}
  \dot{\rho}_{i,j}
  =
  -\frac{\hat{\bm{b}}_{i,j}^{\top}}{\eta_{i,j}}
  \bm{R}_{C_j}^{A}\bm{R}_{B}^{C}\bm{v}_{B,j}.
  \label{eq:rhodot_multi_feature}
\end{equation}
The depth-rate relation isolates the radial contribution of the camera motion.
The remaining part of the relative motion is orthogonal to the bearing
direction and determines the bearing-rate expression used below.

Next, solving \eqref{eq:relative_derivative_static_feature} for
$\dot{\bm{b}}_{i,j}$ using \eqref{eq:rhodot_anchor_velocity} yields
\begin{align}
  \rho_{i,j}\dot{\bm{b}}_{i,j}
  &=
  -{}^{A}\dot{\bm{p}}_{C_j}
  +
  \bm{b}_{i,j}
  \frac{\hat{\bm{b}}_{i,j}^{\top}}{\eta_{i,j}}
  {}^{A}\dot{\bm{p}}_{C_j}.
  \label{eq:bdot_projection_step_1}
\end{align}
Since $\bm{b}_{i,j}=\eta_{i,j}\hat{\bm{b}}_{i,j}$, one has
$\hat{\bm{b}}_{i,j}^{\top}/\eta_{i,j}
=
\bm{b}_{i,j}^{\top}/\eta_{i,j}^{2}$. Therefore,
\begin{align}
  \rho_{i,j}\dot{\bm{b}}_{i,j}
  &=
  -{}^{A}\dot{\bm{p}}_{C_j}
  +
  \frac{1}{\eta_{i,j}^{2}}
  \bm{b}_{i,j}\bm{b}_{i,j}^{\top}
  {}^{A}\dot{\bm{p}}_{C_j} \nonumber\\
  &=
  -\frac{1}{\eta_{i,j}^{2}}
  \left(
  \eta_{i,j}^{2}\bm{I}
  -
  \bm{b}_{i,j}\bm{b}_{i,j}^{\top}
  \right)
  {}^{A}\dot{\bm{p}}_{C_j}.
  \label{eq:bdot_projection_step_2}
\end{align}
Using the projection matrix, $\bm{\pi}_{i,j} \triangleq \eta_{i,j}^{2}\bm{I} -\bm{b}_{i,j}\bm{b}_{i,j}^{\top}$, \eqref{eq:bdot_projection_step_2} becomes
\begin{equation}
  \dot{\bm{b}}_{i,j}
  =
  -\frac{1}{\rho_{i,j}\eta_{i,j}^{2}}
  \bm{\pi}_{i,j}
  {}^{A}\dot{\bm{p}}_{C_j}.
  \label{eq:bdot_anchor_velocity}
\end{equation}
Finally, substituting the translational kinematics
\eqref{eq:pose_velocity} into \eqref{eq:bdot_anchor_velocity} gives the
bearing vector dynamics as
\begin{equation}
  \dot{\bm{b}}_{i,j}
  =
  -\frac{1}{\rho_{i,j}\eta_{i,j}^{2}}
  \bm{\pi}_{i,j}\bm{R}_{C_j}^{A}\bm{R}_{B}^{C}\bm{v}_{B,j}.
  \label{eq:bdot_multi_feature}
\end{equation}

\begin{remark}
Equation~\eqref{eq:bdot_multi_feature} shows that only the component of the
camera translational motion orthogonal to the bearing direction changes the
bearing vector. The projection matrix $\bm{\pi}_{i,j}$ removes the radial
component along $\bm{b}_{i,j}$, whereas the term
$1/(\rho_{i,j}\eta_{i,j}^{2})$ scales the sensitivity of the bearing change.
The rotational quantities enter through the transformation
$\bm{R}_{C_j}^{A}\bm{R}_{B}^{C}$, which maps the body-frame velocity into the
anchor-frame bearing representation.
\end{remark}

The result in \eqref{eq:bdot_multi_feature} is the key intermediate relation
used in the triangulation-aware CBF construction. Since the projection matrix
$\bm{\pi}_{i,j}$ and the stacked-bearing matrix $\bm{M}_i$ are functions of the bearing vectors, the derivation of the bearing vector dynamics provides the path from the commanded translational motion to the derivative of the triangulation-quality barrier. This connection enables the CBF condition to be written in a control-affine form with respect to the current body-frame velocity command.

\subsection{Triangulation-Aware Control Barrier Function Design}
\label{cbfDesign}

The proposed triangulation-aware control barrier function is designed to maintain sufficient aggregate parallax over the active camera-pose window for the tracked features. Rather than treating feature-observability as a passive consequence of the executed trajectory, the proposed formulation embeds a triangulation-quality measure directly into the navigation layer. This requires a triangulation-quality metric whose Lie derivative is control-affine with respect to the current body-frame velocity command, so that corrective translational motion can be synthesized whenever the accumulated bearing geometry approaches a poorly conditioned triangulation configuration. To this end, the feature-wise log-det metric is first defined as
\begin{equation}
  \ell_i(\bm{x}) = \log\det\bm{M}_{i}(\bm{x}),
  \label{eq:ell_i_def}
\end{equation}
where $i = 1,\ldots,s$ denotes the tracked feature index and $s \in \mathbb{N}$ is the number of tracked features. Rather than imposing the CBF condition on a single-feature metric, the proposed formulation uses the average log-det value over all tracked features,
\begin{equation}
  \bar{\ell}(\bm{x})
  = \frac{1}{s}\sum_{i=1}^{s}\ell_i(\bm{x})
  = \frac{1}{s}\sum_{i=1}^{s}\log\det\bm{M}_{i}(\bm{x}).
  \label{eq:average_logdet}
\end{equation}

Equivalently, the average log-det metric can be interpreted as the logarithm of the geometric mean of the feature-wise determinants,
\begin{equation}
  \bar{\ell}(\bm{x})
  =
  \log\left[
  \left(
  \prod_{i=1}^{s}\det\bm{M}_{i}(\bm{x})
  \right)^{1/s}
  \right].
  \label{eq:average_logdet_product}
\end{equation}
This product form shows that the proposed metric evaluates the aggregate triangulation quality over the tracked feature set through the geometric mean of the feature-wise determinant values. Because the barrier is constructed from the average log-det value, the resulting condition lower-bounds aggregate triangulation quality rather than imposing an individual lower bound on every feature. Let $\bar{\ell}_{\min}\in\R$ denote the prescribed minimum acceptable average log-det value. The barrier function is then defined as
\begin{align}
    h(\bm{x})
    &= \bar{\ell}(\bm{x}) - \bar{\ell}_{\min} \nonumber\\
    &=
    \log\left[
    \left(
    \prod_{i=1}^{s}\det\bm{M}_{i}(\bm{x})
    \right)^{1/s}
    \right]
    - \bar{\ell}_{\min}.
  \label{eq:h_average_def}
\end{align}

Thus, the safe set becomes
\begin{align}
  \mathcal{C}
  &= \left\{\bm{x}: h(\bm{x})\ge0\right\} \nonumber\\
  &=
  \left\{\bm{x}:
  \log\left[
  \left(
  \prod_{i=1}^{s}\det\bm{M}_{i}(\bm{x})
  \right)^{1/s}
  \right]
  \ge \bar{\ell}_{\min}
  \right\}.
  \label{eq:safe_set_average}
\end{align}

Then, the barrier derivative can be evaluated as
\begin{equation}
  \dot{h}(\bm{x})
  =
  \frac{1}{s}\sum_{i=1}^{s}\dot{\ell}_{i}.
  \label{eq:hdot_average_metric}
\end{equation}

For each positive-definite $\bm{M}_{i}$,
\begin{equation}
  \dot{\ell}_{i}
  = \frac{d}{dt}\log\det\bm{M}_{i}
  = \Tr\left(\bm{M}_{i}^{-1}\dot{\bm{M}}_{i}\right).
  \label{eq:ellidot_logdet_identity}
\end{equation}
Using \eqref{eq:Mi_def},
\begin{align}
  \dot{\bm{M}}_{i}
  &= \sum_{j=1}^{n}\dot{\bm{\pi}}_{i,j} \nonumber\\
  &= -\sum_{j=1}^{n}
  \left(\dot{\bm{b}}_{i,j}\bm{b}_{i,j}^{\top}
  + \bm{b}_{i,j}\dot{\bm{b}}_{i,j}^{\top}\right),
  \label{eq:Midot}
\end{align}
where $\eta_{i,j}$ is assumed constant for normalized bearing measurements. For unit-normalized bearings, $\eta_{i,j}=1$. Substituting \eqref{eq:Midot} into \eqref{eq:ellidot_logdet_identity} gives
\begin{align}
  \dot{\ell}_{i}
  &=
  -\sum_{j=1}^{n}
  \Tr\left[
  \bm{M}_{i}^{-1}
  \left(
  \dot{\bm{b}}_{i,j}\bm{b}_{i,j}^{\top}
  +
  \bm{b}_{i,j}\dot{\bm{b}}_{i,j}^{\top}
  \right)
  \right] \nonumber\\
  &=
  -\sum_{j=1}^{n}
  \left[
  \Tr\left(
  \bm{M}_{i}^{-1}\dot{\bm{b}}_{i,j}\bm{b}_{i,j}^{\top}
  \right)
  +
  \Tr\left(
  \bm{M}_{i}^{-1}\bm{b}_{i,j}\dot{\bm{b}}_{i,j}^{\top}
  \right)
  \right] \nonumber\\
  &=
  -\sum_{j=1}^{n}
  \left[
  \bm{b}_{i,j}^{\top}\bm{M}_{i}^{-1}\dot{\bm{b}}_{i,j}
  +
  \dot{\bm{b}}_{i,j}^{\top}\bm{M}_{i}^{-1}\bm{b}_{i,j}
  \right].
  \label{eq:ellidot_trace_expanded}
\end{align}
Since $\bm{M}_{i}^{-1}$ is symmetric, the two scalar terms in \eqref{eq:ellidot_trace_expanded} are equal. Therefore,
\begin{equation}
  \dot{\ell}_{i}
  = -2\sum_{j=1}^{n}
  \bm{b}_{i,j}^{\top}\bm{M}_{i}^{-1}\dot{\bm{b}}_{i,j}.
  \label{eq:ellidot_compact}
\end{equation}

Using the bearing-rate model in \eqref{eq:bdot_multi_feature}, the feature-wise metric derivative becomes
\begin{align}
  \dot{\ell}_{i}
  &=
  -2\sum_{j=1}^{n}
  \bm{b}_{i,j}^{\top}\bm{M}_{i}^{-1}
  \left(
  -\frac{1}{\rho_{i,j}\eta_{i,j}^{2}}
  \bm{\pi}_{i,j}\bm{R}_{C_j}^{A}\bm{R}_{B}^{C}\bm{v}_{B,j}
  \right) \nonumber\\
  &=
  2\sum_{j=1}^{n}
  \frac{1}{\rho_{i,j}\eta_{i,j}^{2}}
  \bm{b}_{i,j}^{\top}\bm{M}_{i}^{-1}\bm{\pi}_{i,j}
  \bm{R}_{C_j}^{A}\bm{R}_{B}^{C}\bm{v}_{B,j}.
  \label{eq:ellidot_after_bdot}
\end{align}
Since $\dot{h}(\bm{x}) = (1/s)\sum_{i=1}^{s}\dot{\ell}_{i}$, substituting \eqref{eq:ellidot_after_bdot} into \eqref{eq:hdot_average_metric} yields
\begin{align}
  \dot{h}(\bm{x})
  &= \frac{2}{s}\sum_{i=1}^{s}\sum_{j=1}^{n}
  \frac{1}{\rho_{i,j}\eta_{i,j}^{2}}
  \bm{b}_{i,j}^{\top}\bm{M}_{i}^{-1}\bm{\pi}_{i,j}
  \bm{R}_{C_j}^{A}\bm{R}_{B}^{C}\bm{v}_{B,j}.
  \label{eq:hdot_average_expanded}
\end{align}

Here, the active camera-pose window is treated as a time-indexed trajectory segment; hence, the bearing-rate contributions associated with previous poses are evaluated from recorded velocity data, whereas only the current-pose velocity remains designable. Because only the current-pose velocity $\bm{v}_{B,n}\equiv\bm{v}_{B}$ is the control input while all past-pose velocities $\bm{v}_{B,j}$, $j<n$, are recorded quantities within the active camera-pose window, the double sum splits naturally into a drift term and a control term. To isolate the designable control input, \eqref{eq:hdot_average_expanded} is separated into the past-pose contributions $j=1,\ldots,n-1$ and the current-pose contribution $j=n$ as
\begin{align}
  \dot{h}(\bm{x})
  &=
  \frac{2}{s}\sum_{i=1}^{s}\sum_{j=1}^{n-1}
  \frac{1}{\rho_{i,j}\eta_{i,j}^{2}}
  \bm{b}_{i,j}^{\top}\bm{M}_{i}^{-1}\bm{\pi}_{i,j}
  \bm{R}_{C_j}^{A}\bm{R}_{B}^{C}\bm{v}_{B,j}
  \nonumber\\
  &\quad+
  \frac{2}{s}\sum_{i=1}^{s}
  \frac{1}{\rho_{i,n}\eta_{i,n}^{2}}
  \bm{b}_{i,n}^{\top}\bm{M}_{i}^{-1}\bm{\pi}_{i,n}
  \bm{R}_{C_n}^{A}\bm{R}_{B}^{C}\bm{v}_{B}.
  \label{eq:hdot_split_current_past}
\end{align}
For the current-pose term, the scalar identity
\begin{align}
  &\bm{b}_{i,n}^{\top}\bm{M}_{i}^{-1}\bm{\pi}_{i,n}
  \bm{R}_{C_n}^{A}\bm{R}_{B}^{C}\bm{v}_{B} \nonumber\\
  &\quad =
  \left[
  \left(\bm{R}_{C_n}^{A}\bm{R}_{B}^{C}\right)^{\top}
  \bm{\pi}_{i,n}\bm{M}_{i}^{-1}\bm{b}_{i,n}
  \right]^{\top}
  \bm{v}_{B}
  \label{eq:current_term_scalar_identity}
\end{align}
is used, where the symmetry of $\bm{M}_{i}^{-1}$ and $\bm{\pi}_{i,n}$ have been exploited. Therefore, the barrier derivative can be written in the control-affine form
\begin{equation}
  \dot{h}(\bm{x})
  =
  \nabla_{\bm{x}} h(\bm{x})^{\top}\dot{\bm{x}}
  =
  \underbrace{
  \overbrace{\mathcal{L}_{\bm{F}}h(\bm{x})}^{\triangleq\, f(\bm{x})}
  }_{\text{drift}}
  +
  \underbrace{
  \overbrace{\mathcal{L}_{\bm{G}}h(\bm{x})}^{\triangleq\, \bm{g}(\bm{x})^{\top}}
  }_{\text{control}}
  \bm{v}_{B},
  \label{eq:hdot_drift_control}
\end{equation}
where
\begin{align}
  f(\bm{x})
  &= \frac{2}{s}\sum_{i=1}^{s}\sum_{j=1}^{n-1}
  \frac{1}{\rho_{i,j}\eta_{i,j}^{2}}
  \bm{b}_{i,j}^{\top}\bm{M}_{i}^{-1}\bm{\pi}_{i,j}
  \bm{R}_{C_j}^{A}\bm{R}_{B}^{C}\bm{v}_{B,j},
  \label{eq:drift_def}\\[6pt]
  \bm{g}(\bm{x})
  &= \frac{2}{s}\sum_{i=1}^{s}
  \frac{1}{\rho_{i,n}\eta_{i,n}^{2}}
  \left(\bm{R}_{C_n}^{A}\bm{R}_{B}^{C}\right)^{\top}
  \bm{\pi}_{i,n}\bm{M}_{i}^{-1}\bm{b}_{i,n}.
  \label{eq:g_def}
\end{align}

From the decomposition in \eqref{eq:hdot_drift_control}, the barrier derivative has the control-affine structure with the drift and control-input gradient defined in \eqref{eq:drift_def}--\eqref{eq:g_def}. The scalar drift comprises the known contributions of the recorded past-pose velocities and is therefore computable at each time step from the trajectory window. Furthermore, $\bm{g}(\bm{x})$ is the body-frame gradient that multiplies the designable velocity command.

The compact average log-det CBF condition is then imposed using the linear extended class-$\mathcal{K}_{\infty}$ function $\alpha(h)=\gamma h$, with $\gamma \in \mathbb{R}_{>0}$, as
\begin{equation}
  \begin{aligned}
    &f(\bm{x})+\bm{g}(\bm{x})^{\top}\bm{v}_{B}
    +\gamma
    \left(
    \log\left[
    \left(
    \prod_{i=1}^{s}\det\bm{M}_{i}(\bm{x})
    \right)^{1/s}
    \right]
    - \bar{\ell}_{\min}
    \right)\\
    &\hspace{\fill}\ge 0.
  \end{aligned}
  \label{eq:cbf_condition_average}
\end{equation}
Equivalently, by the definition of $h(\bm{x})$ in \eqref{eq:h_average_def},
\begin{equation}
  f(\bm{x})+\bm{g}(\bm{x})^{\top}\bm{v}_{B}
  +\gamma h(\bm{x}) \ge 0.
  \label{eq:cbf_condition_compact}
\end{equation}
If the resulting command satisfies \eqref{eq:cbf_condition_compact} for all $t \geq 0$, then
\begin{equation}
  \dot{h}(\bm{x}) \ge -\alpha(h(\bm{x})).
\end{equation}
Since $\alpha(\cdot)$ is selected as an extended class-$\mathcal{K}_{\infty}$ function, the condition is well-defined on both sides of the boundary $h(\bm{x})=0$. In particular, for any trajectory initialized in the safe set, i.e., $h(\bm{x}(0))\ge0$, the comparison lemma implies that $h(\bm{x}(t))\ge0$ for all $t\ge0$. Hence, the aggregate triangulation-quality safe set $\mathcal{C}$ is forward invariant whenever the CBF condition is imposed as a hard constraint and the corresponding admissible velocity set is nonempty.

\subsection{Mission-Dependent Refinement}

The hard CBF condition in \eqref{eq:cbf_condition_compact} provides the strictest form of triangulation-quality maintenance. However, practical navigation tasks may also involve higher-priority mission requirements, such as stopping, hovering, actuator saturation, or preserving a prescribed trajectory. In these cases, enforcing the triangulation-aware CBF as an unconditional hard constraint may require a velocity command that is far from the nominal command in magnitude, or it may conflict with the mission objective itself. To obtain a formulation that can represent both strict and relaxed operation, the CBF condition is written in a triangulation- and mission-priority-aware form by introducing a nonnegative slack variable $\delta_T\in\R_{\ge0}$:
\begin{equation}
  f(\bm{x})+\bm{g}(\bm{x})^{\top}\bm{v}_{B}
  +\gamma h(\bm{x})+\delta_T \ge 0.
  \label{eq:soft_cbf_condition}
\end{equation}
The slack variable $\delta_T\in\mathbb{R}_{\ge0}$ quantifies the instantaneous relaxation of the triangulation-maintenance constraint and allows the same formulation to represent two operating choices. When strict enforcement of the prescribed triangulation-quality threshold is required, $\delta_T$ is fixed to zero, and \eqref{eq:soft_cbf_condition} reduces exactly to the hard CBF condition in \eqref{eq:cbf_condition_compact}. When the nominal command represents a higher-priority mission objective, $\delta_T$ is optimized, allowing controlled relaxation of the strict forward-invariance guarantee associated with the prescribed threshold. Thus, the slack variable is not introduced merely as a numerical device; rather, it acts as a priority variable whose value directly indicates the instantaneous conflict between triangulation-quality maintenance and nominal mission tracking.

Finally, the CBF-QP setup is formulated using a weighted deviation cost.
Let $\bm{W}_{B}\in\R^{3\times3}$ denote a symmetric positive-definite weighting matrix, i.e., $\bm{W}_{B}=\bm{W}_{B}^{\top}$ and $\bm{a}^{\top}\bm{W}_{B}\bm{a}>0$ $\forall \bm{a}\in\R^{3}\setminus\{\bm{0}\}$. The weighting matrix is introduced to shape the geometry of the velocity correction. In an unweighted minimum-deviation filter, the correction is aligned with the Euclidean projection direction induced by the CBF gradient $\bm{g}(\bm{x})$, which may rotate the corrected velocity command away from the intended direction of motion. By contrast, the weighted formulation allows deviations along the instantaneous nominal direction to be penalized less than deviations in the perpendicular subspace, thereby favoring speed adjustment before lateral redirection whenever the nominal direction is well-defined. Let the instantaneous nominal body-frame direction be defined as
\begin{equation}
  \hat{\bm{n}}_{B}
  =
  \frac{\bm{v}_{B,\mathrm{nom}}}
  {\|\bm{v}_{B,\mathrm{nom}}\|_{2}},
  \label{eq:nominal_tangent_direction}
\end{equation}
where $\|\bm{v}_{B,\mathrm{nom}}\|_{2}>\varepsilon_v$ and $\varepsilon_v>0$ is a small threshold used to avoid direction ambiguity at very low nominal speeds. When the nominal speed falls below this threshold, no instantaneous tangential direction is assigned and the correction should not be interpreted as direction preserving; in that case, the active CBF correction is governed by the barrier gradient and the selected positive-definite weighting. The tangent and perpendicular projection matrices are then defined as
\begin{align}
  \bm{P}_{\parallel} &= \hat{\bm{n}}_{B}\hat{\bm{n}}_{B}^{\top}, \\
  \bm{P}_{\perp} &= \bm{I}-\hat{\bm{n}}_{B}\hat{\bm{n}}_{B}^{\top},
  \label{eq:tangent_perpendicular_projection}
\end{align}
where $\bm{P}_{\parallel},\bm{P}_{\perp}\in\R^{3\times3}$. The weighting
matrix is selected as
\begin{equation}
  \bm{W}_{B}
  =
  w_{\parallel}\bm{P}_{\parallel}
  +
  w_{\perp}\bm{P}_{\perp},
  \label{eq:weighted_metric_matrix}
\end{equation}
where $w_{\parallel}$ and $w_{\perp}$ denote the tangent and perpendicular
weights, respectively and satisfy $0<w_{\parallel}<w_{\perp}$. Since $\bm{P}_{\parallel}$ and $\bm{P}_{\perp}$ are orthogonal projection matrices satisfying $\bm{P}_{\parallel}+\bm{P}_{\perp}=\bm{I}$, the matrix $\bm{W}_{B}$ is symmetric positive definite for $0<w_{\parallel}<w_{\perp}$. Hence, $\bm{W}_{B}^{-1}$ exists. Moreover, because $w_{\parallel}<w_{\perp}$, deviations parallel to the current nominal velocity are penalized less than deviations in the perpendicular subspace. Therefore, when the CBF constraint becomes active and the nominal velocity is nonzero, the safety filter preferentially modifies the speed along the intended direction of motion before introducing lateral velocity components. This interpretation is not applied during terminal stop or hover intervals, where the nominal tangential direction is not well-defined.

Using the weighted metric and the relaxed CBF condition, the triangulation- and priority-aware TANGO-VIO safety filter is formulated as
\begin{equation}
  \begin{aligned}
  \bm{v}_{B}^{*},\delta_T^{*}
  =
  \underset{\substack{\bm{v}_{B}\in\R^{3}\\\delta_T\in\R}}{\operatorname{arg\,min}}
  &\quad
  \frac{1}{2}
  \left\|
  \bm{v}_{B}-\bm{v}_{B,\mathrm{nom}}
  \right\|_{\bm{W}_{B}}^{2}
  +
  \frac{p_T}{2}\delta_T^{2}\\
  \mathrm{s.t.}
  &\quad
  f(\bm{x})+\bm{g}(\bm{x})^{\top}\bm{v}_{B}
  +\gamma h(\bm{x})+\delta_T\ge0,\\
  &\quad
  \delta_T\ge0.
  \end{aligned}
  \label{eq:soft_weighted_cbf_qp}
\end{equation}
where $p_T>0$ is the triangulation-priority weight. A larger value of $p_T$ penalizes CBF relaxation more strongly, so the optimizer behaves closer to the hard-CBF formulation. A smaller value of $p_T$ allows more relaxation of the triangulation constraint, thereby keeping the corrected velocity command closer to the nominal command when the two objectives conflict. The hard-CBF formulation is recovered from \eqref{eq:soft_weighted_cbf_qp} by imposing $\delta_T=0$.

To derive the closed-form solution of \eqref{eq:soft_weighted_cbf_qp}, define the CBF residual evaluated at the nominal command as
\begin{equation}
  \mu \triangleq f(\bm{x}) + \bm{g}(\bm{x})^{\top}\bm{v}_{B,\mathrm{nom}} + \gamma h(\bm{x}).
  \label{eq:soft_mu_def}
\end{equation}
If $\mu\ge0$, the nominal command already satisfies the hard CBF condition. Therefore, the optimal solution is
\begin{equation}
  \bm{v}_{B}^{*}=\bm{v}_{B,\mathrm{nom}},
  \qquad
  \delta_T^{*}=0.
  \label{eq:soft_inactive_solution}
\end{equation}

For the active case $\mu<0$, the relaxed CBF constraint is active at the optimum. The Lagrangian of \eqref{eq:soft_weighted_cbf_qp} is written directly in terms of $\bm{v}_{B}$ as
\begin{align}
  \mathfrak{L}(\bm{v}_{B},\delta_T,\lambda)
  =&
  \frac{1}{2}
  \left\|
  \bm{v}_{B}-\bm{v}_{B,\mathrm{nom}}
  \right\|_{\bm{W}_{B}}^{2}
  +
  \frac{p_T}{2}\delta_T^{2} \nonumber\\
  &-
  \lambda
  \left(
  f(\bm{x})+\bm{g}(\bm{x})^{\top}\bm{v}_{B}
  +\gamma h(\bm{x})+\delta_T
  \right),
  \label{eq:soft_lagrangian}
\end{align}
where $\lambda\in\R_{\ge0}$ is the Lagrange multiplier associated with the relaxed CBF constraint. The stationarity condition with respect to $\bm{v}_{B}$ gives
\begin{equation}
  \frac{\partial\mathfrak{L}}{\partial \bm{v}_{B}}
  =
  \bm{W}_{B}\left(\bm{v}_{B}-\bm{v}_{B,\mathrm{nom}}\right)
  -
  \lambda\bm{g}(\bm{x})
  =
  \bm{0}.
  \label{eq:soft_stationarity_v}
\end{equation}
Since $\bm{W}_{B}$ is invertible, \eqref{eq:soft_stationarity_v} yields
\begin{equation}
  \bm{v}_{B}^{*}
  =
  \bm{v}_{B,\mathrm{nom}}
  +
  \lambda^{*}\bm{W}_{B}^{-1}\bm{g}(\bm{x}).
  \label{eq:soft_v_solution_lambda}
\end{equation}
The stationarity condition with respect to $\delta_T$ gives
\begin{equation}
  \frac{\partial\mathfrak{L}}{\partial \delta_T}
  =
  p_T\delta_T-\lambda
  =
  0.
  \label{eq:soft_stationarity_delta}
\end{equation}
Therefore,
\begin{equation}
  \delta_T^{*}
  =
  \frac{\lambda^{*}}{p_T}.
  \label{eq:soft_delta_solution}
\end{equation}
Substituting \eqref{eq:soft_v_solution_lambda} and \eqref{eq:soft_delta_solution} into the active CBF constraint gives
\begin{align}
  0
  &=
  f(\bm{x})
  +
  \bm{g}(\bm{x})^{\top}
  \left(
  \bm{v}_{B,\mathrm{nom}}
  +
  \lambda^{*}\bm{W}_{B}^{-1}\bm{g}(\bm{x})
  \right)
  +
  \gamma h(\bm{x})
  +
  \frac{\lambda^{*}}{p_T} \nonumber\\
  &=
  \mu
  +
  \lambda^{*}
  \left(
  \bm{g}(\bm{x})^{\top}\bm{W}_{B}^{-1}\bm{g}(\bm{x})
  +
  \frac{1}{p_T}
  \right).
  \label{eq:soft_lambda_substitution}
\end{align}
Define $S(\bm{x}) \triangleq \bm{g}(\bm{x})^{\top}\bm{W}_{B}^{-1}\bm{g}(\bm{x})$. The optimal multiplier is obtained as
\begin{equation}
  \lambda^{*}
  =
  -
  \frac{\mu}
  {S(\bm{x})+\frac{1}{p_T}}.
  \label{eq:soft_lambda_solution}
\end{equation}
Since the active case occurs only when $\mu<0$, the multiplier satisfies $\lambda^{*}>0$, which is consistent with the Karush-Kuhn-Tucker (KKT) condition $\lambda^{*}\ge0$.

Consequently, the closed-form velocity command is
\begin{equation}
  \bm{v}_{B}^{*}
  =
  \begin{cases}
  \bm{v}_{B,\mathrm{nom}},
  & \mu\ge0,\\[8pt]
  \bm{v}_{B,\mathrm{nom}}
  -
  \dfrac{\mu}
  {S(\bm{x})+\dfrac{1}{p_T}}
  \bm{W}_{B}^{-1}\bm{g}(\bm{x}),
  & \mu<0,
  \end{cases}
  \label{eq:soft_closed_form_velocity}
\end{equation}
and the optimal slack is
\begin{equation}
  \delta_T^{*}
  =
  \begin{cases}
  0,
  & \mu\ge0,\\[8pt]
  \dfrac{-\mu}
  {1+p_T S(\bm{x})},
  & \mu<0.
  \end{cases}
  \label{eq:soft_closed_form_slack}
\end{equation}

Equations~\eqref{eq:soft_closed_form_velocity}--\eqref{eq:soft_closed_form_slack} show that the optimizer does not simply minimize the slack variable alone. Instead, it minimizes the combined cost of velocity-command deviation and triangulation-constraint relaxation. For any selected velocity command, the smallest admissible slack is used; however, the globally optimal solution may intentionally use a nonzero slack if satisfying the hard CBF would require an excessive deviation from $\bm{v}_{B,\mathrm{nom}}$. 

The relaxed CBF condition modifies the theoretical guarantee. From \eqref{eq:soft_cbf_condition}, one obtains
\begin{equation}
  \dot{h}(\bm{x})
  \ge
  -\gamma h(\bm{x})-\delta_T.
  \label{eq:soft_cbf_differential_inequality}
\end{equation}
Therefore, the original forward-invariance guarantee is preserved when $\delta_T(t)=0$ for all $t$. When $\delta_T(t)>0$, the triangulation-quality constraint is intentionally relaxed, and $\delta_T(t)$ quantifies the instantaneous loss of the strict CBF guarantee.

\section{Results}
\label{results}

The proposed TANGO-VIO layer is evaluated in both software-in-the-loop (SITL) simulations and
flight-test experiments. In both cases, visual-inertial odometry is provided by
OpenVINS~\cite{geneva2020openvins}, an open-source filter-based MSCKF-SLAM 
pipeline, which serves as one representative instantiation of the sliding-window
structure described in Section~\ref{slidingWindowFeatureTracks}. 
OpenVINS distinguishes between short-track MSCKF features, which are
triangulated and discarded once they leave the window, and persistent
(SLAM) features, which are maintained in the state vector and re-observed across the window; in this implementation, the persistent (SLAM) features constitute the multi-view tracked feature set on which the proposed barrier is evaluated. The SITL analysis considers square and climbing trajectories in nominal, hard-CBF, and soft-CBF modes, whereas the flight campaign considers only the climbing trajectory in nominal and hard-CBF modes. The results are analyzed through the commanded-motion response of the proposed layer and the resulting visual-information quality.

\subsection{Software-in-the-Loop Results}
\label{SITL}

\subsubsection{SITL Setup and Baseline Definition}

\begin{figure*}[!tb]
    \centering
    \includegraphics[width=\textwidth]{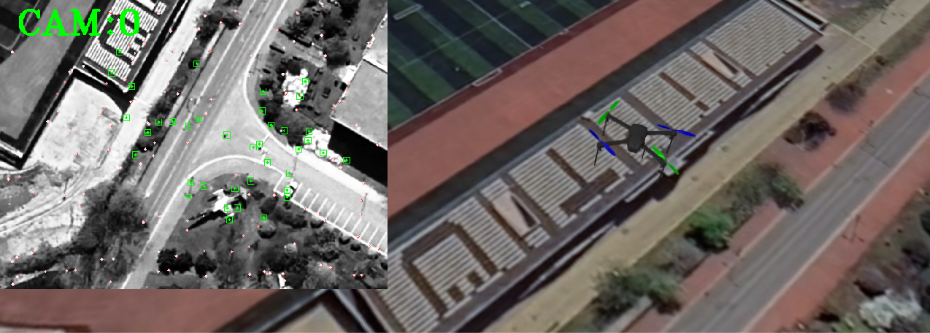}
    \caption{Software-in-the-loop environment used for the repeatable evaluation of TANGO-VIO. The background shows the flat map implemented in Gazebo, while the inset in the upper-left corner presents the camera view processed by the VIO front end, with green boxes indicating the triangulated persistent (SLAM) features.}
    \label{fig:sitl_view_group}
\end{figure*}

The SITL setup combines a Gazebo environment with camera and IMU sensor models, the ArduPilot SITL flight stack, and ROS~2 communication through MAVROS, as shown in Fig.~\ref{fig:sitl_view_group}. At each update, the active camera-pose window and tracked feature bearings are used to compute the stacked-bearing matrices, average log-det barrier, and minimum-deviation body-frame velocity command. The analytical filter was run at $20~\mathrm{Hz}$ with $\gamma=3$, $w_{\parallel}=1$, $w_{\perp}=10$, $\bar{\ell}_{\min}=-2.5$, and a $10~\mathrm{m/s}$ velocity-command limit; $p_T=5$ was used in the soft-CBF cases. This rate was sufficient for the experiments, while the closed-form implementation permits higher update rates. The threshold was selected empirically through preliminary trials, and no general tuning rule is proposed.

Two representative scenarios are considered.

\begin{scenario}[Square trajectory]
    The vehicle follows a square path with $100~\mathrm{m}$ side length at an altitude of $60~\mathrm{m}$ and an average nominal speed of $3~\mathrm{m/s}$. The start and finish points coincide, so the nominal mission includes a terminal stop at the initial position.
\end{scenario}

\begin{scenario}[Climbing trajectory]
    The vehicle follows a forward-climbing path with $3~\mathrm{m/s}$ nominal body Forward speed and $1~\mathrm{m/s}$ nominal body Up speed, increasing the altitude from $30~\mathrm{m}$ until it traverses $100~\mathrm{m}$ distance in the forward direction.
\end{scenario}

For each scenario, three operating modes are compared. In the nominal baseline, the TANGO-VIO correction is disabled and the commanded velocity is the reference velocity generated by the B-Spline based mission planner. In the hard-CBF mode, the triangulation constraint in \eqref{eq:cbf_condition_compact} is imposed without relaxation; therefore, maintaining the prescribed log-det threshold has priority over terminal stopping or exact path tracking. In the soft-CBF mode, the slack-enabled formulation in \eqref{eq:soft_cbf_condition}--\eqref{eq:soft_closed_form_slack} is used, allowing the controller to trade instantaneous triangulation-quality enforcement against deviation from the nominal command.

\subsubsection{CBF-Induced Navigation Response}
\label{cbf_induced_navigation_response}

The commanded-motion response of the proposed layer is evaluated using the barrier-constraint evolution, the deviation between the nominal and corrected velocity commands, and the resulting trajectory-level motion.

\begin{figure*}[!tb]
    \centering
    \subfloat[Velocity command modification.\label{fig:square_hard_velocityComparison}]{%
        \includegraphics[width=0.32\textwidth]{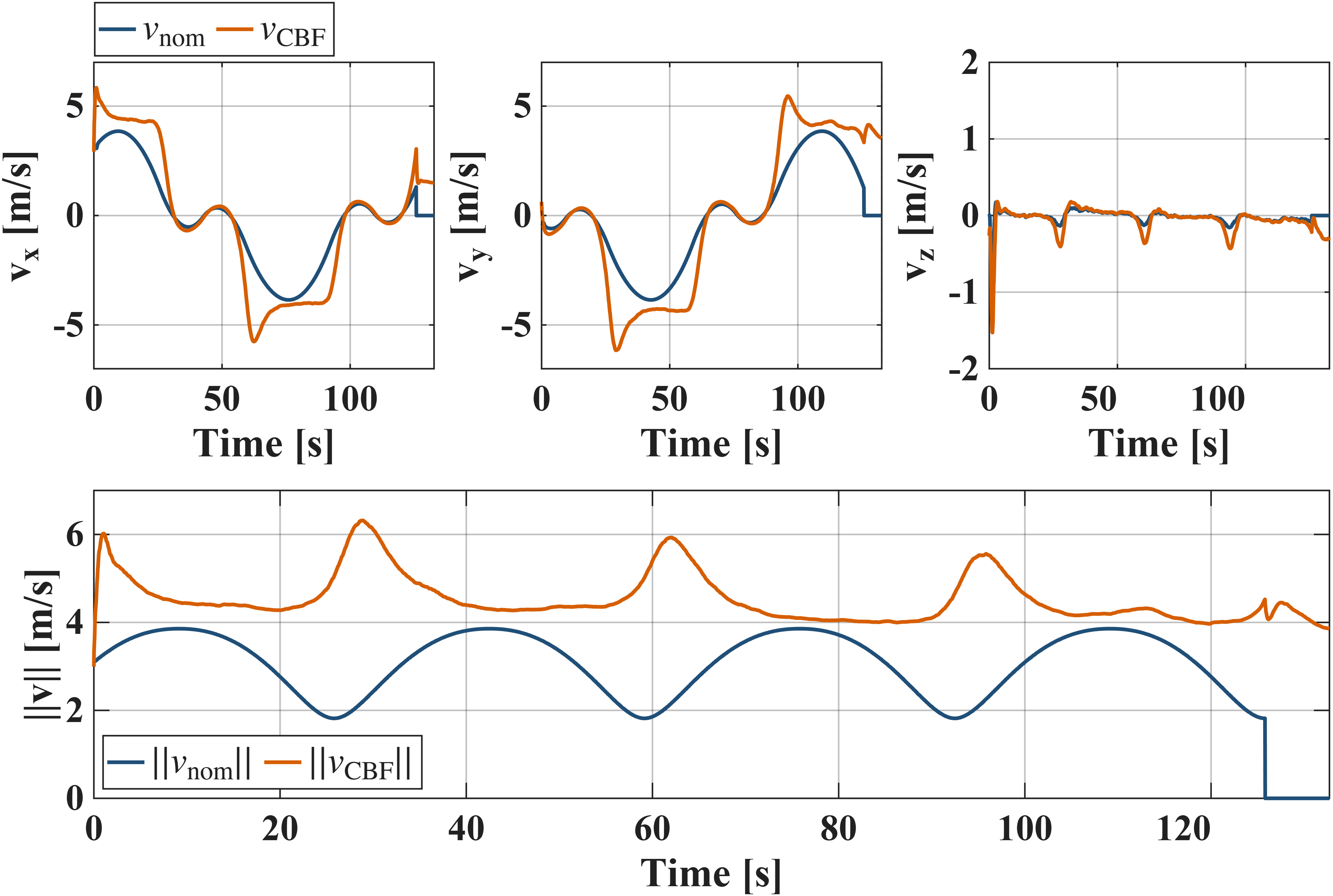}}
    \hfill
    \subfloat[Barrier-constraint history.\label{fig:square_hard_CBFHistory}]{%
        \includegraphics[width=0.32\textwidth]{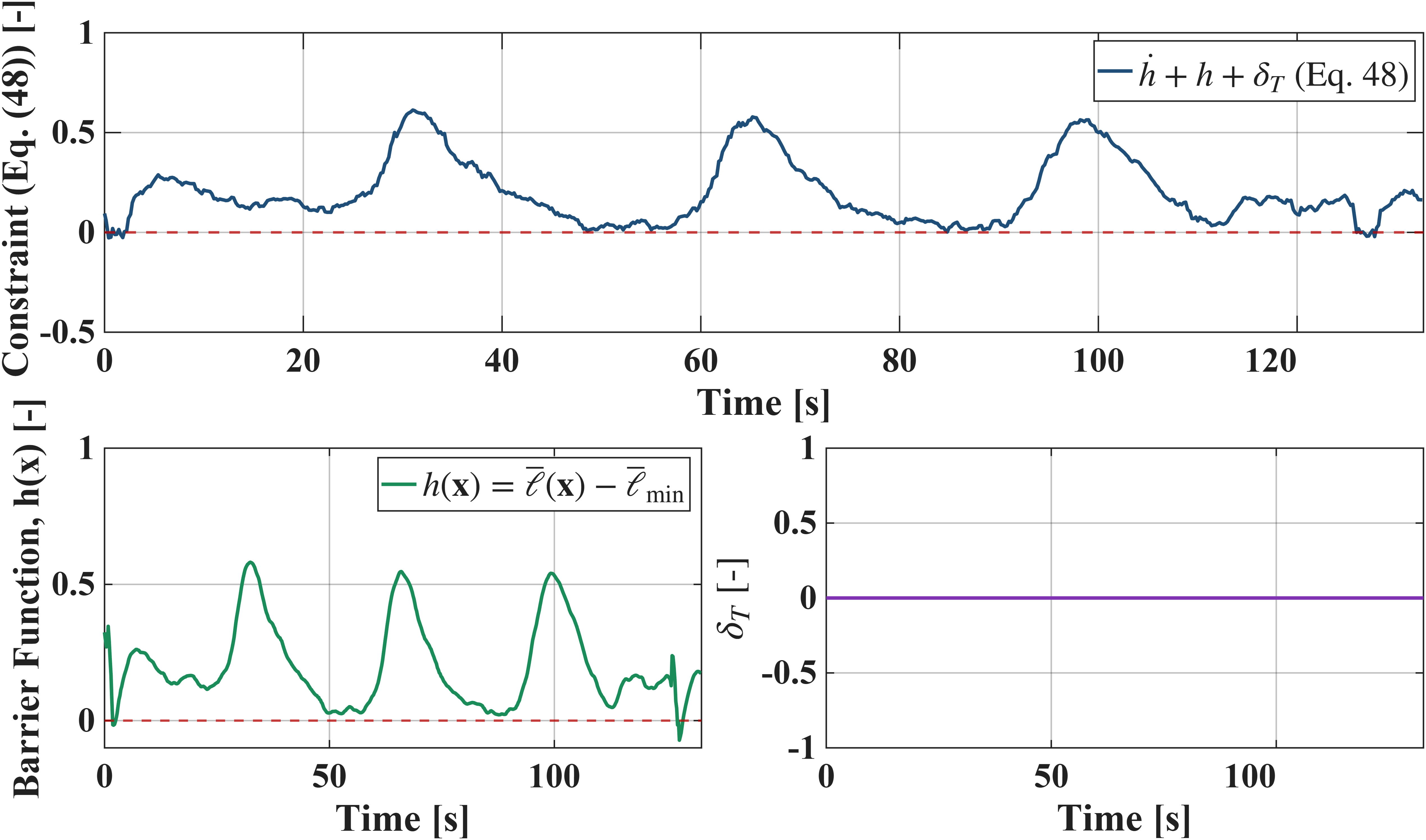}}
    \hfill
    \subfloat[Three-dimensional response.\label{fig:square_hard_3Dtraj}]{%
        \includegraphics[width=0.32\textwidth]{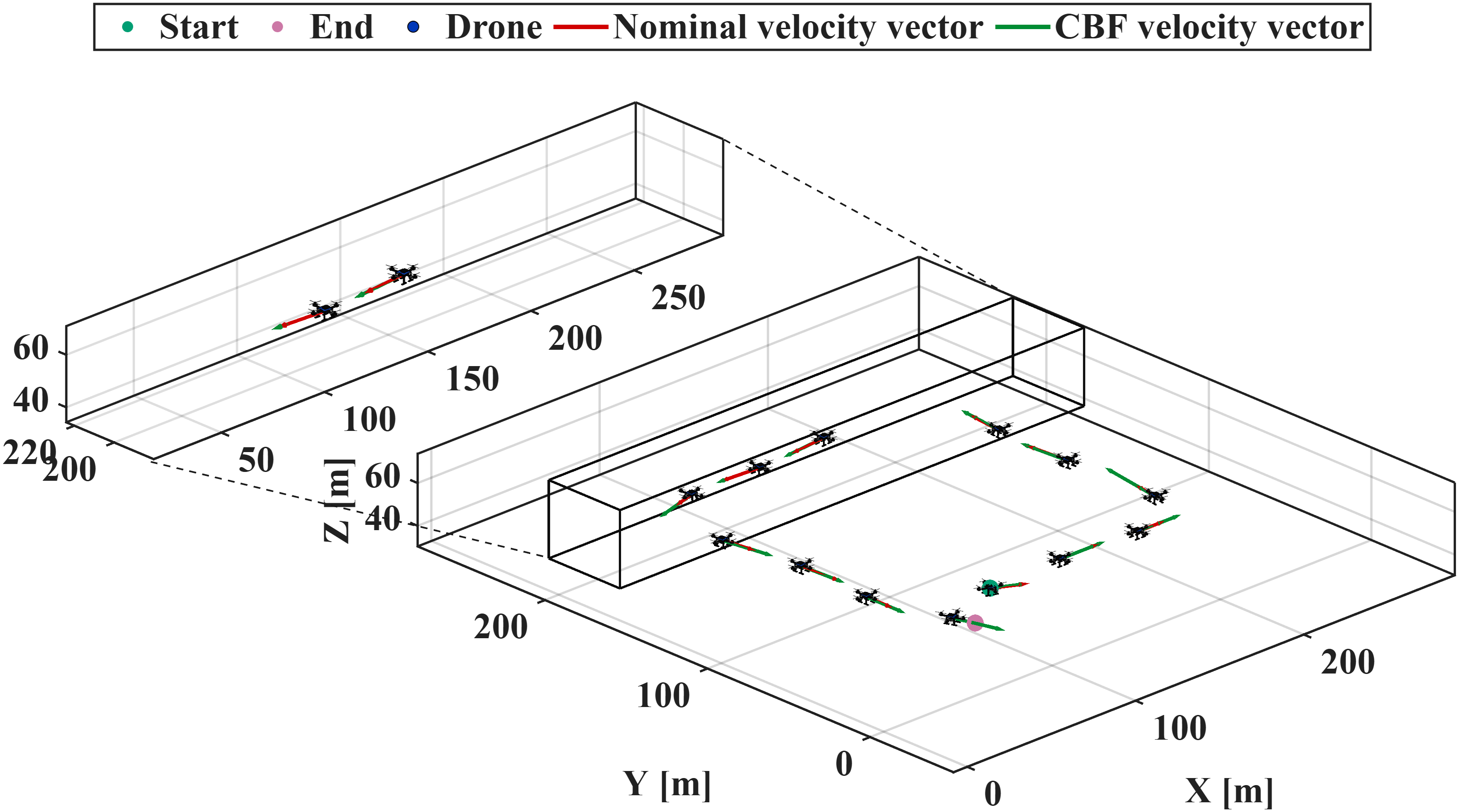}}
    \caption{CBF-induced navigation response for the square trajectory under the hard constraint. The unrelaxed triangulation constraint produces velocity corrections when the nominal command becomes insufficient for maintaining parallax, leading to the larger trajectory-level deviation from the nominal square.}
    \label{fig:square_hard_cbf_response}
\end{figure*}

\emph{Square trajectory:}
For the square trajectory, the hard-CBF formulation produces a strict triangulation-priority response. As shown in Fig.~\ref{fig:square_hard_cbf_response}, the corrected command departs from the nominal velocity whenever the barrier condition approaches activity. The barrier history verifies satisfaction of the unrelaxed CBF inequality, while the three-dimensional trajectory shows the corresponding geometric consequence: the path expands relative to the nominal square and the vehicle is not allowed to stop exactly at the nominal terminal point because doing so would remove the translational excitation needed for triangulation.

\begin{figure*}[!tb]
    \centering
    \subfloat[Velocity command modification.\label{fig:square_soft_velocityComparison}]{%
        \includegraphics[width=0.32\textwidth]{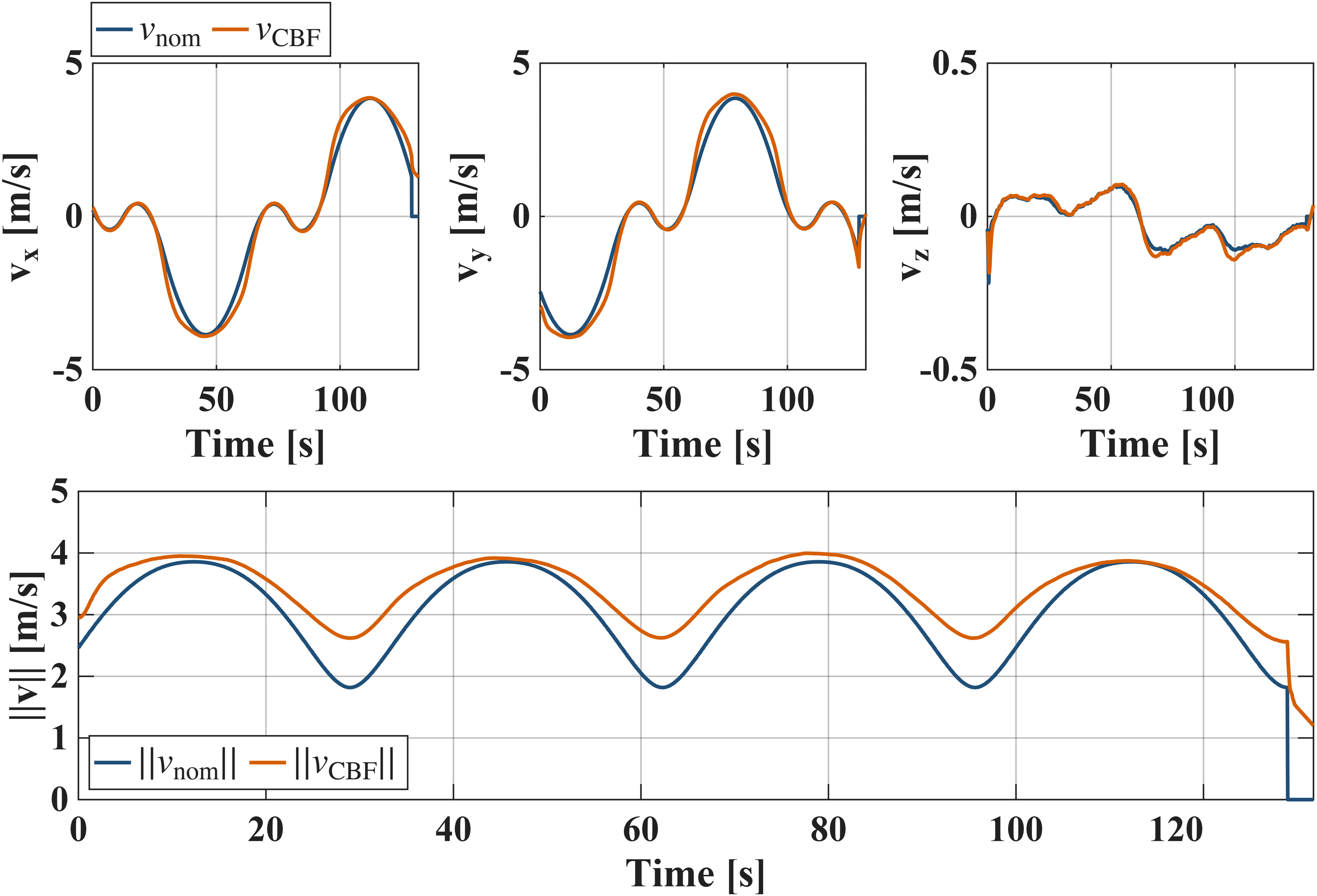}}
    \hfill
    \subfloat[Relaxed barrier and slack history.\label{fig:square_soft_CBFHistory}]{%
        \includegraphics[width=0.32\textwidth]{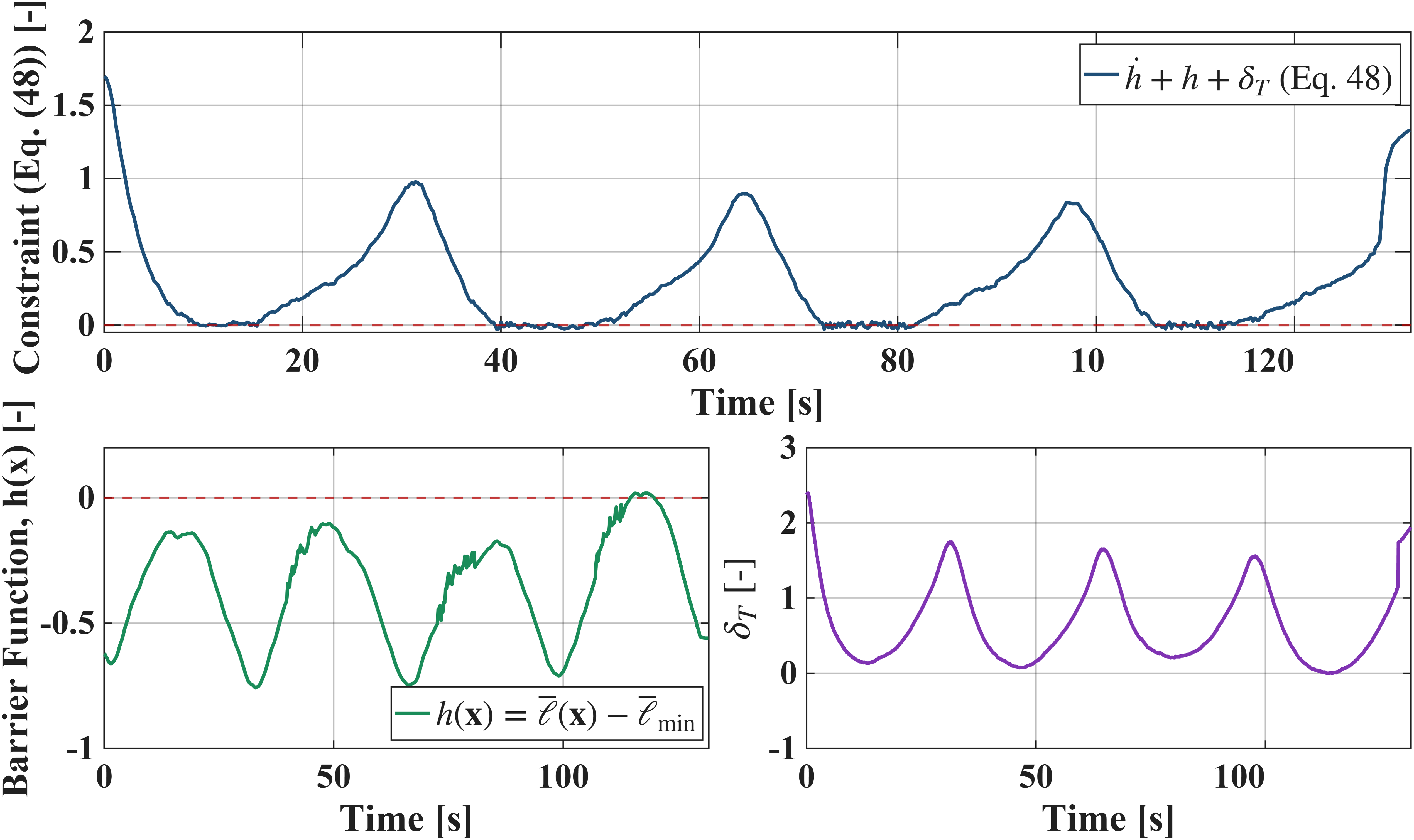}}
    \hfill
    \subfloat[Three-dimensional response.\label{fig:square_soft_3Dtraj}]{%
        \includegraphics[width=0.32\textwidth]{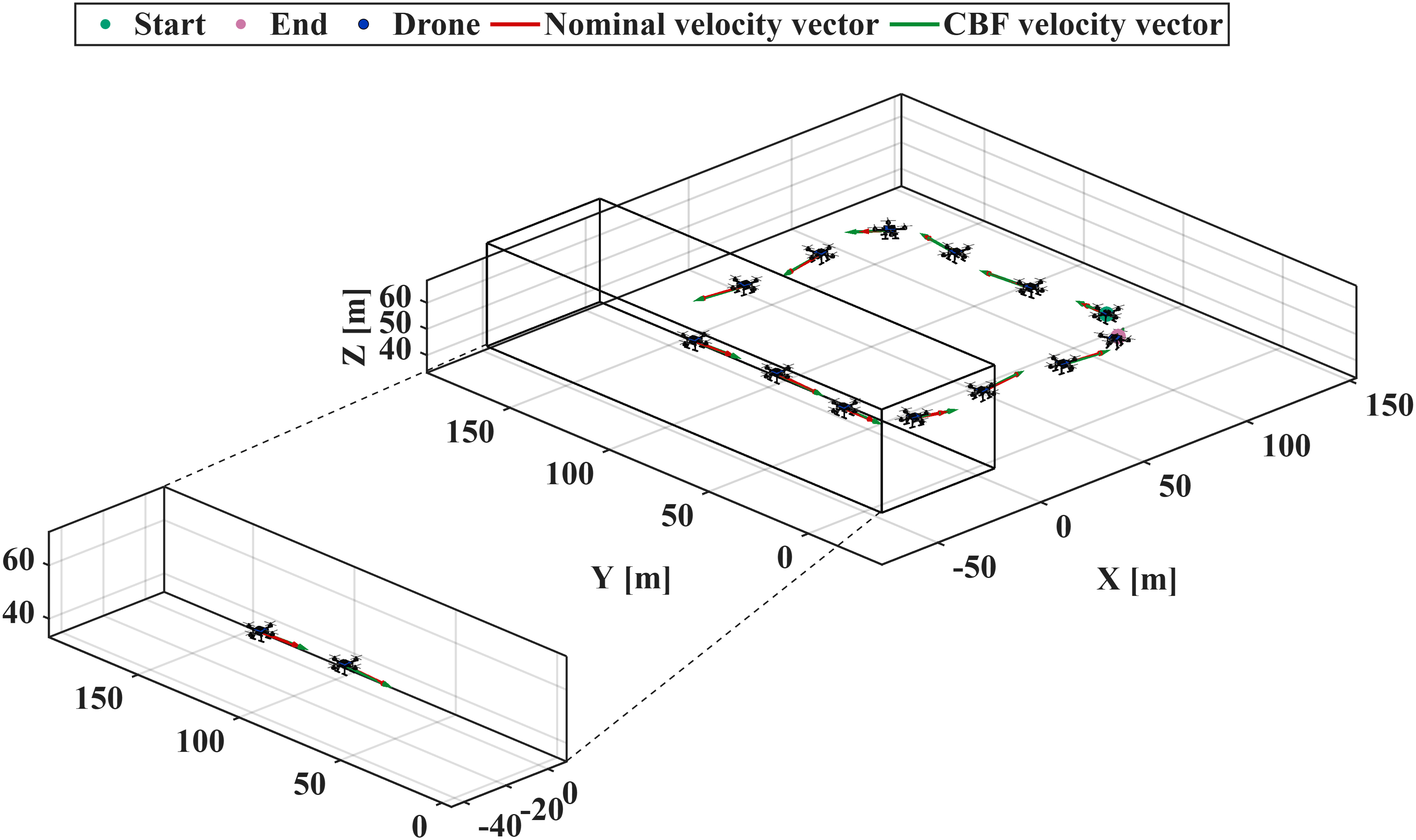}}
    \caption{CBF-induced navigation response for the square trajectory under the soft constraint. The slack-enabled formulation keeps the corrected command closer to the nominal command than the hard-CBF case while retaining triangulation-aware corrections in poorly conditioned visual-geometry intervals.}
    \label{fig:square_soft_cbf_response}
\end{figure*}

The soft-CBF result in Fig.~\ref{fig:square_soft_cbf_response} exhibits the same corrective mechanism with reduced intervention severity. The corrected velocity remains closer to the nominal command, particularly near the terminal portion of the mission. The slack history identifies the instants at which strict enforcement of the original log-det threshold competes with the nominal path-following objective. Consequently, the soft-CBF trajectory retains the nominal square geometry more closely while still introducing parallax-enhancing motion in the low-excitation intervals.

\begin{figure}[!t]
    \centering
    \includegraphics[width=\columnwidth]{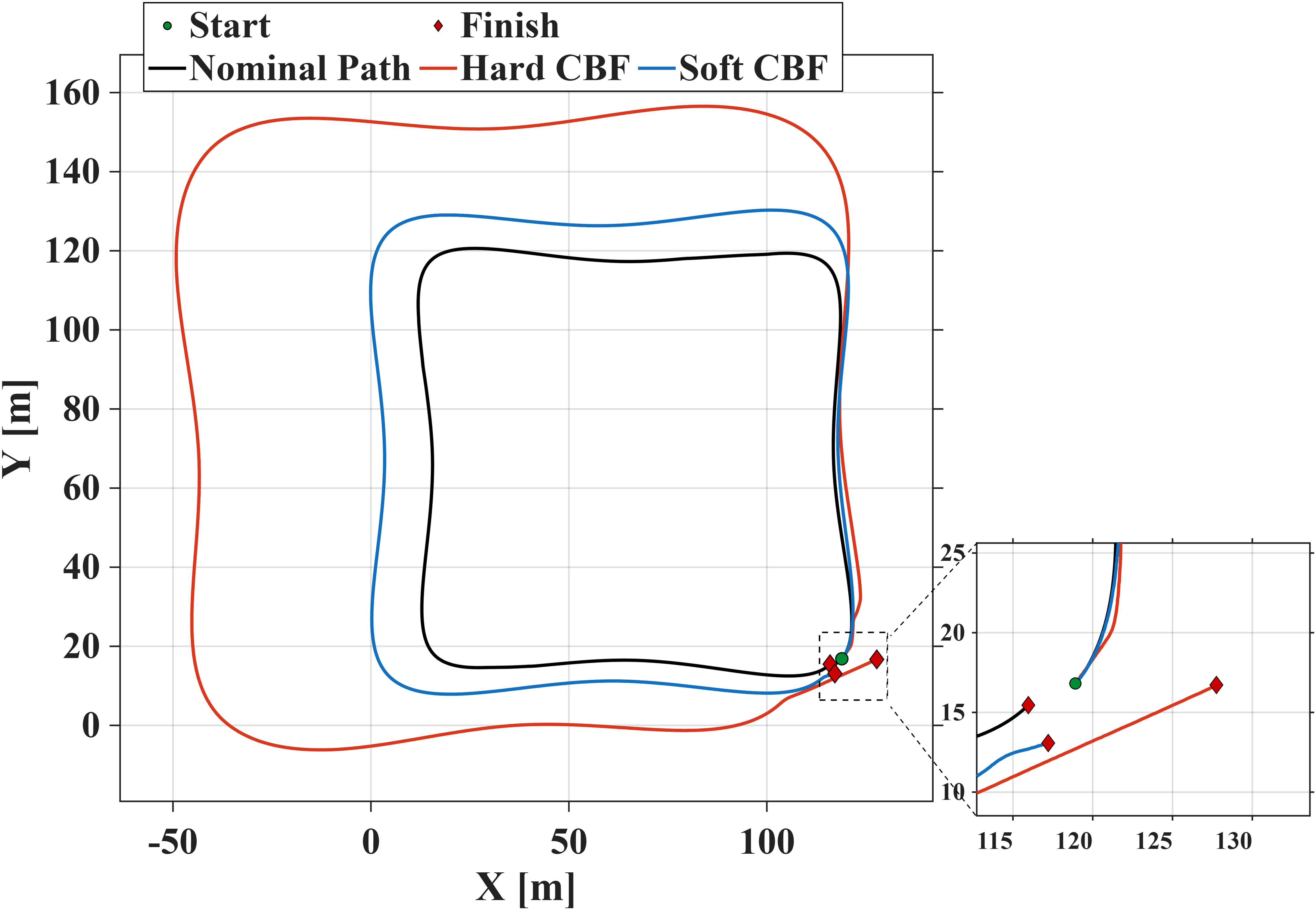}
    \caption{Trajectory-level comparison for the square scenario. The hard CBF produces the larger path distortion and shifts the terminal behavior, whereas the soft CBF preserves the main shape of the nominal trajectory more closely by allowing controlled relaxation of the triangulation constraint.}
    \label{fig:square_trajComparison}
\end{figure}

The aggregate trajectory comparison in Fig.~\ref{fig:square_trajComparison} summarizes the effect of the two constraint choices. The hard-CBF trajectory expands substantially because triangulation preservation is treated as a strict requirement. The soft-CBF trajectory remains much closer to the nominal path, showing that slack-enabled operation reduces path distortion while retaining the ability to react to low-parallax motion.

\begin{figure*}[!tb]
    \centering
    \subfloat[Velocity command modification.\label{fig:climb_hard_velocityComparison}]{%
        \includegraphics[width=0.32\textwidth]{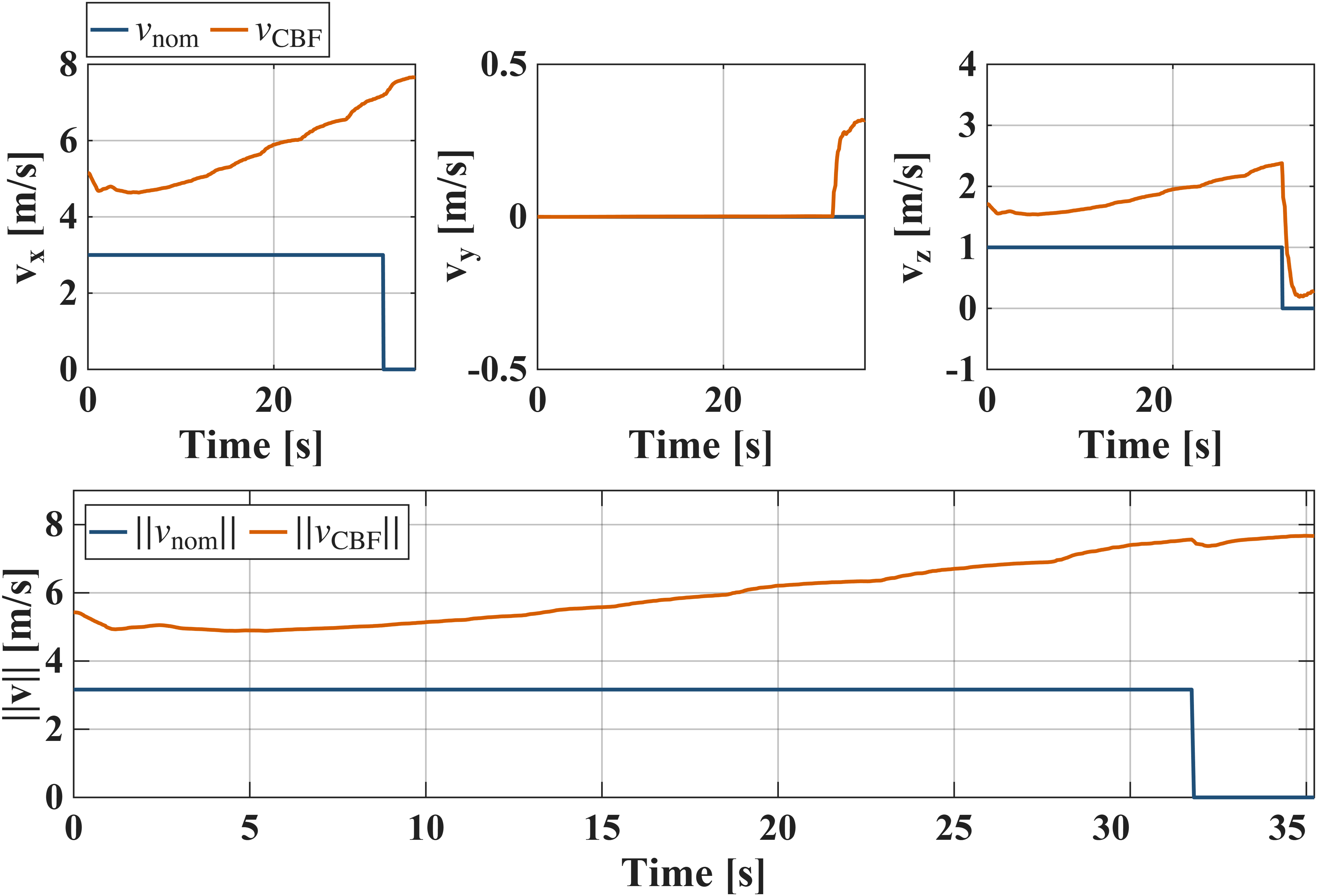}}
    \hfill
    \subfloat[Barrier-constraint history.\label{fig:climb_hard_CBFHistory}]{%
        \includegraphics[width=0.32\textwidth]{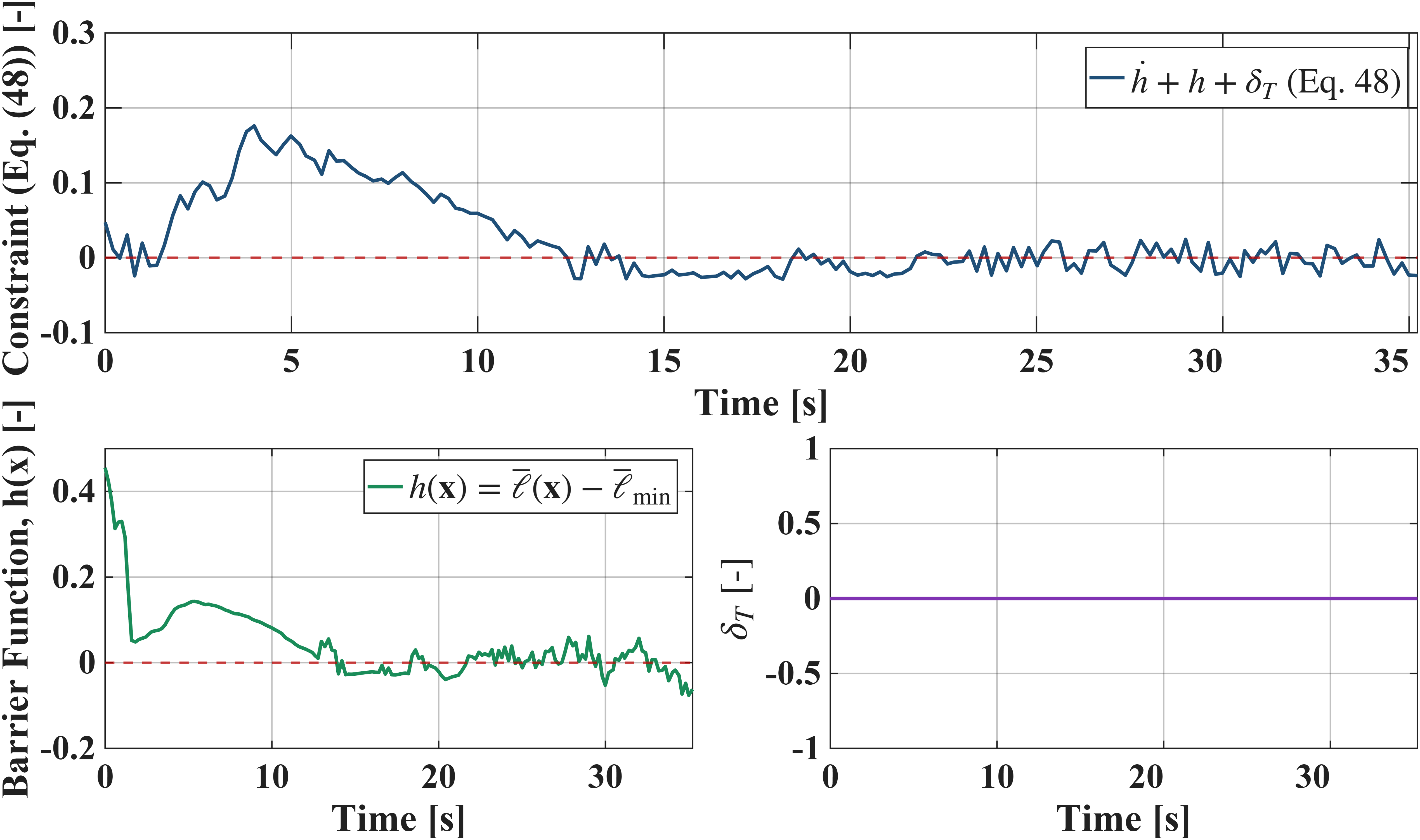}}
    \hfill
    \subfloat[Three-dimensional response.\label{fig:climb_hard_3Dtraj}]{%
        \includegraphics[width=0.32\textwidth]{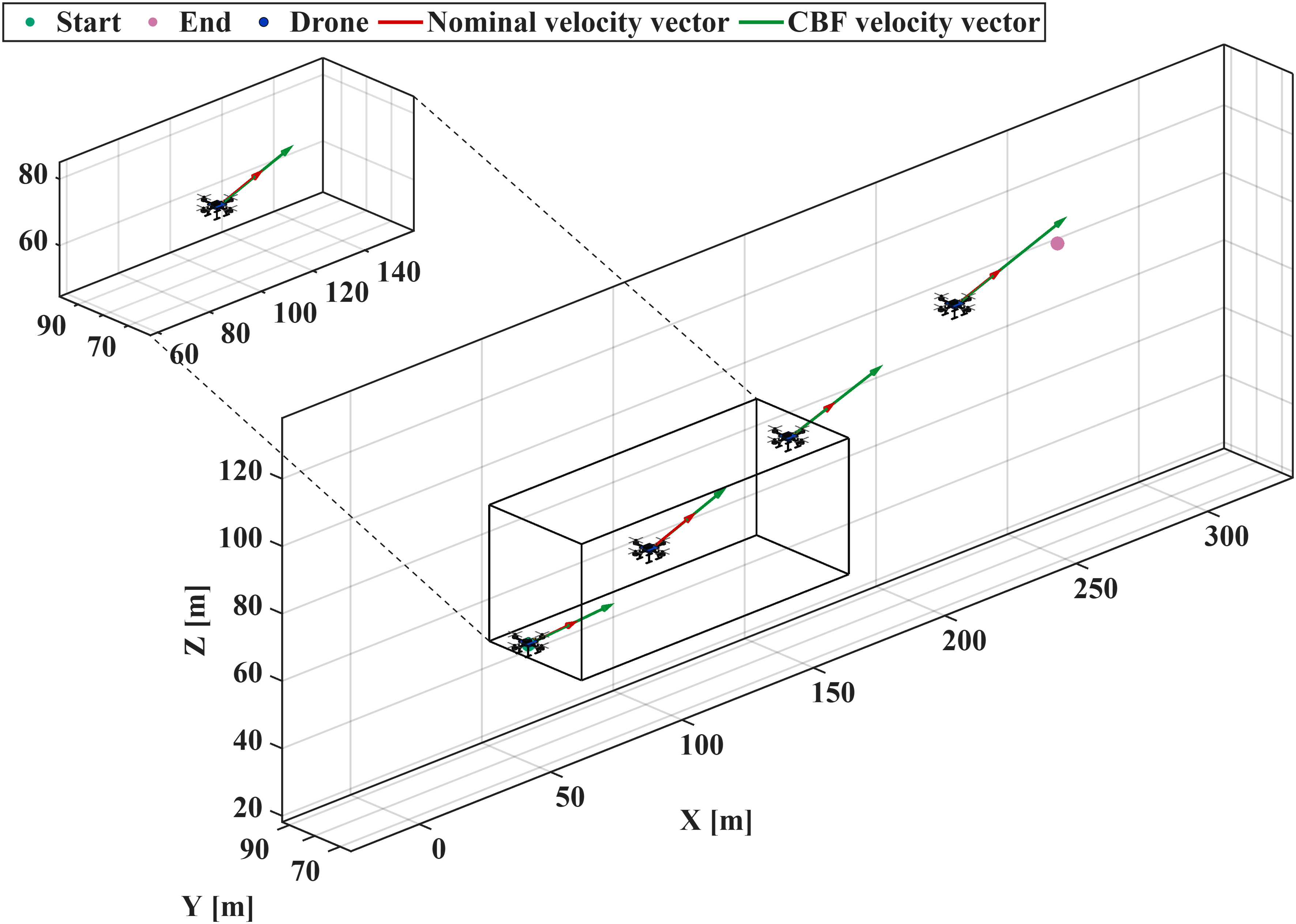}}
    \caption{CBF-induced navigation response for the climbing trajectory under the hard constraint. The strict triangulation-priority command introduces additional motion when the nominal climb does not provide sufficient parallax, producing terminal deviation from the nominal path.}
    \label{fig:climb_hard_cbf_response}
\end{figure*}

\emph{Climbing trajectory:}
The climbing trajectory evaluates the proposed layer under coupled forward and vertical motion. Under the hard constraint, Fig.~\ref{fig:climb_hard_cbf_response} shows that the CBF filter modifies the forward and vertical velocity components when the nominal climb becomes insufficient to maintain feature-observability. The barrier history confirms satisfaction of the unrelaxed condition, and the three-dimensional trajectory illustrates the resulting terminal deviation. This behavior is not only a consequence of the terminal low-velocity phase; as altitude increases, the translational motion required to maintain well-defined triangulation also increases. Therefore, a constant nominal velocity can become insufficient after a certain altitude even before the terminal segment is reached.

\begin{figure*}[!tb]
    \centering
    \subfloat[Velocity command modification.\label{fig:climb_soft_velocityComparison}]{%
        \includegraphics[width=0.32\textwidth]{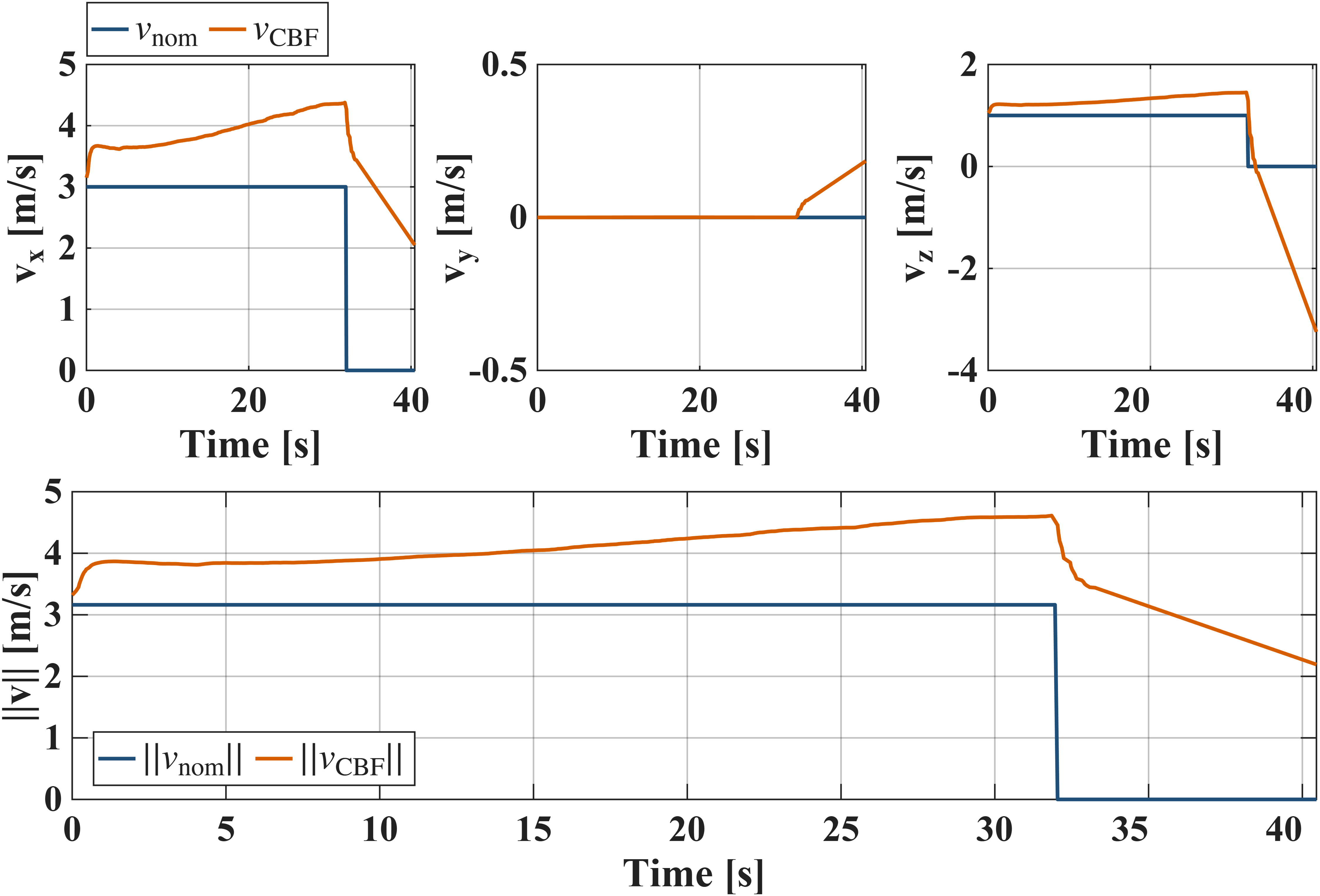}}
    \hfill
    \subfloat[Relaxed barrier and slack history.\label{fig:climb_soft_CBFHistory}]{%
        \includegraphics[width=0.32\textwidth]{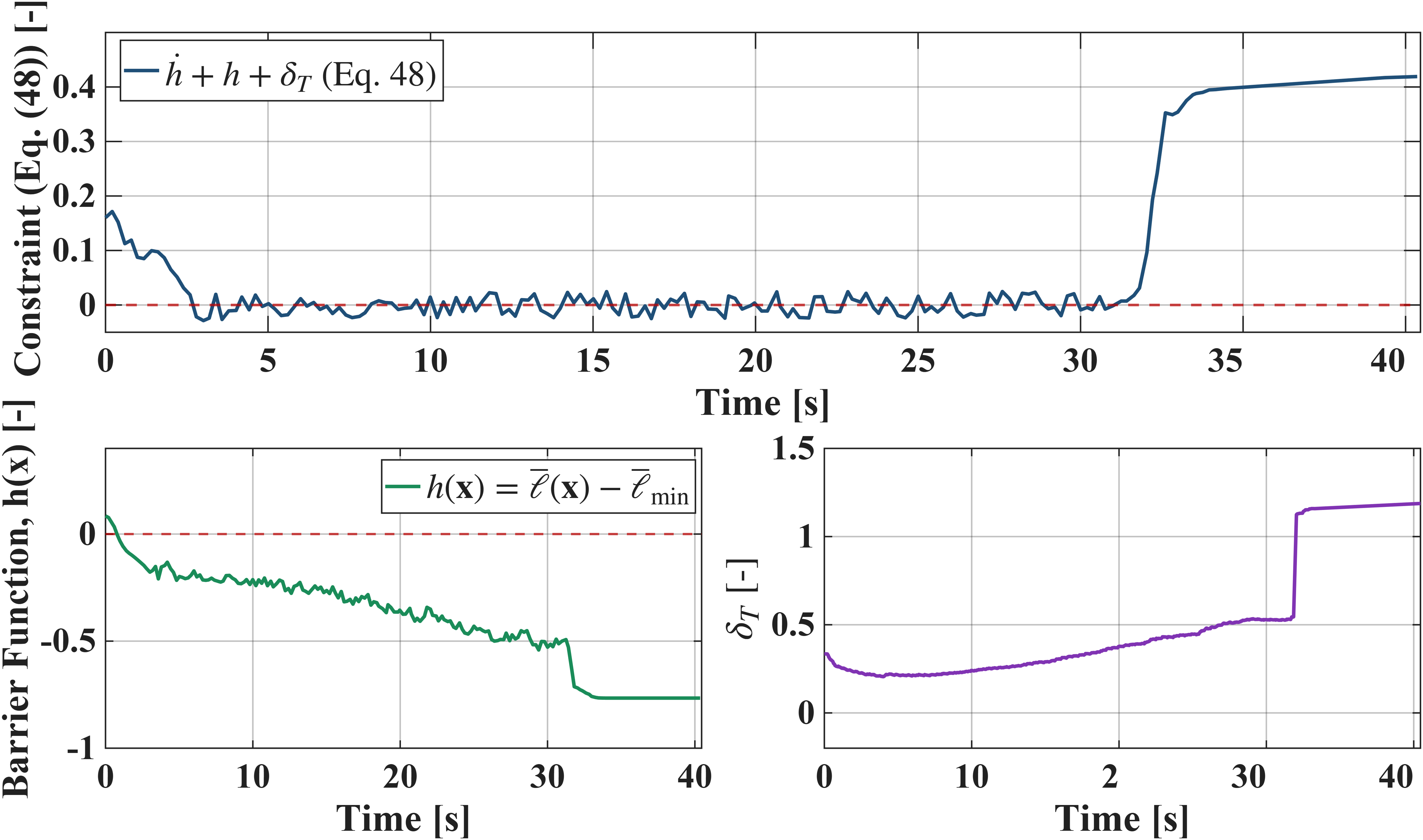}}
    \hfill
    \subfloat[Three-dimensional response.\label{fig:climb_soft_3Dtraj}]{%
        \includegraphics[width=0.32\textwidth]{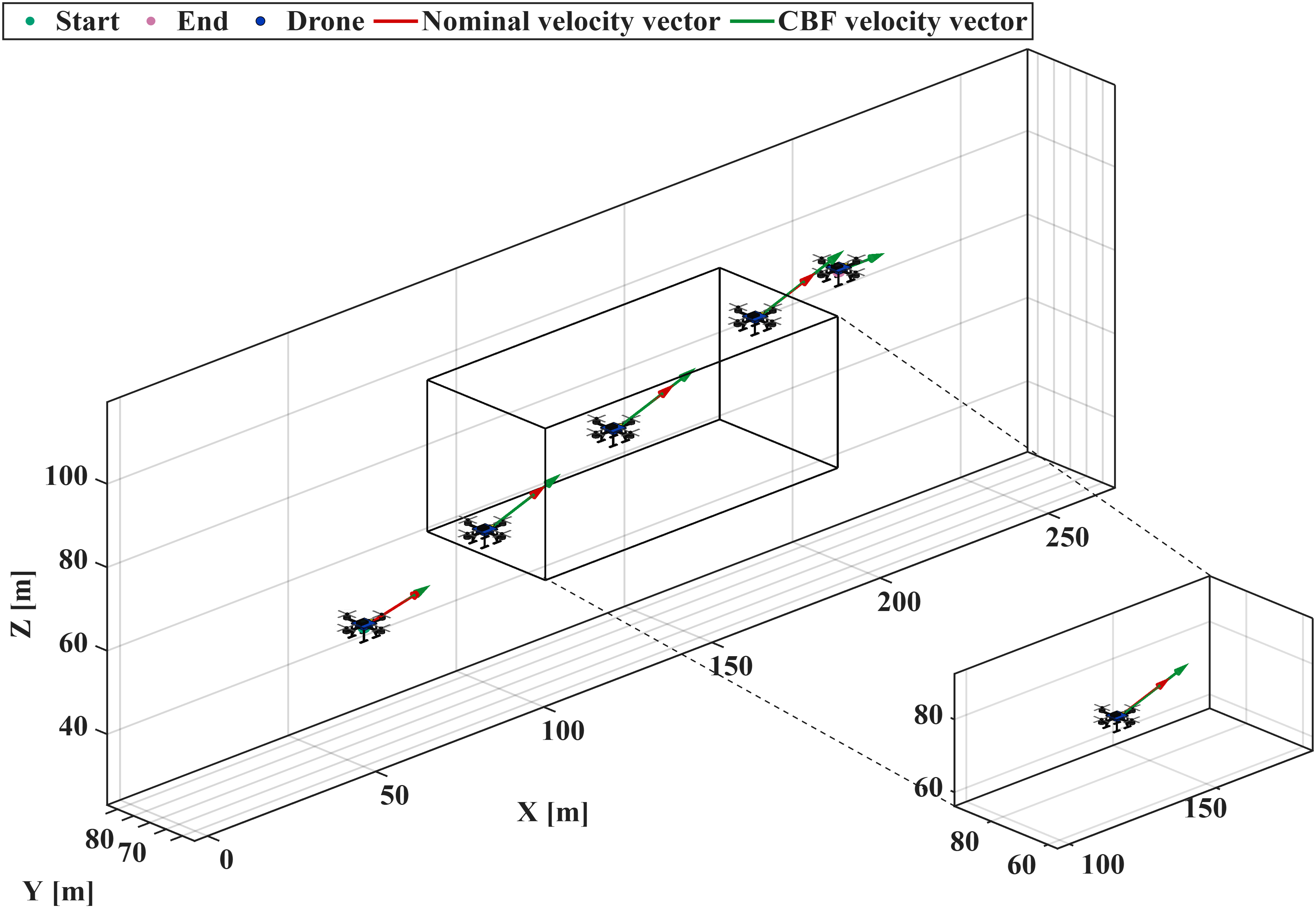}}
    \caption{CBF-induced navigation response for the climbing trajectory under the soft constraint. The slack-enabled correction remains closer to the nominal command and reduces the terminal deviation observed with the hard constraint. Near the terminal stop, the nominal tangential direction is not well-defined, and the soft formulation may favor slack relaxation over continued parallax-generating motion.}
    \label{fig:climb_soft_cbf_response}
\end{figure*}

With the soft constraint, Fig.~\ref{fig:climb_soft_cbf_response} shows a less aggressive but still active response. The velocity correction remains closer to the nominal command than in the hard-CBF case, while the slack history records the temporary relaxation needed to balance terminal path-following against triangulation-quality maintenance. In the terminal portion, the soft formulation can permit the vehicle to slow down or stop instead of continuously injecting translational excitation. The three-dimensional trajectory therefore exhibits reduced terminal deviation, making the soft formulation more appropriate when the mission requires both visual-information support and preservation of the prescribed terminal behavior.

\begin{figure}[!t]
    \centering
    \includegraphics[width=\columnwidth]{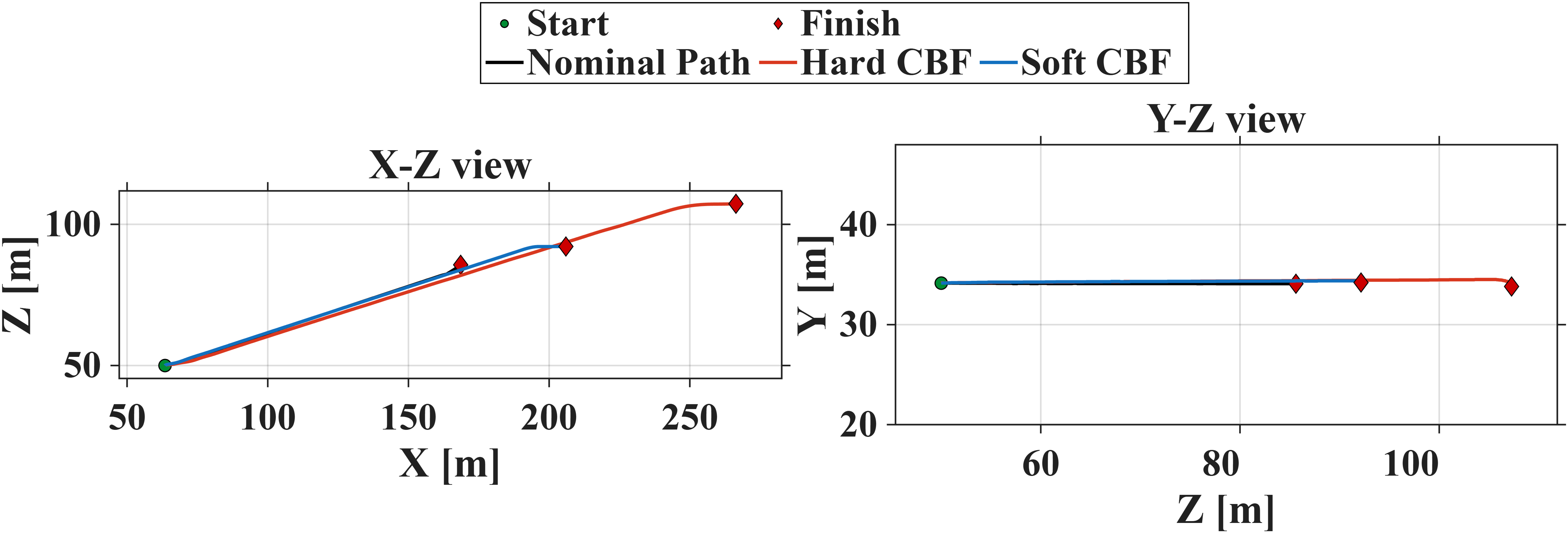}
    \caption{Trajectory-level comparison for the climbing scenario. The hard CBF preserves the strict triangulation constraint at the expense of terminal path tracking, whereas the soft CBF follows the nominal climb more closely by allowing limited relaxation through the slack variable.}
    \label{fig:climb_trajComparison}
\end{figure}

The aggregate comparison in Fig.~\ref{fig:climb_trajComparison} confirms the same tradeoff observed in the square scenario. The hard CBF maintains the strict triangulation constraint at the expense of terminal path tracking, whereas the soft CBF significantly reduces this deviation by permitting controlled relaxation when the mission objective requires the vehicle to slow down or stop.

\subsubsection{Resulting Visual-Information Quality}
\label{resulting_visual_information_quality}

The effect of the CBF-induced motion on the visual front end is assessed using the number of newly triangulated features and the total persistent-feature count (SLAM features in OpenVINS terminology). These quantities characterize whether the commanded motion supplies sufficient visual baseline for continuous feature initialization and tracking.

\begin{figure*}[!tb]
    \centering
    \subfloat[Nominal.\label{fig:square_disabled_featureCount}]{%
        \includegraphics[width=0.32\textwidth]{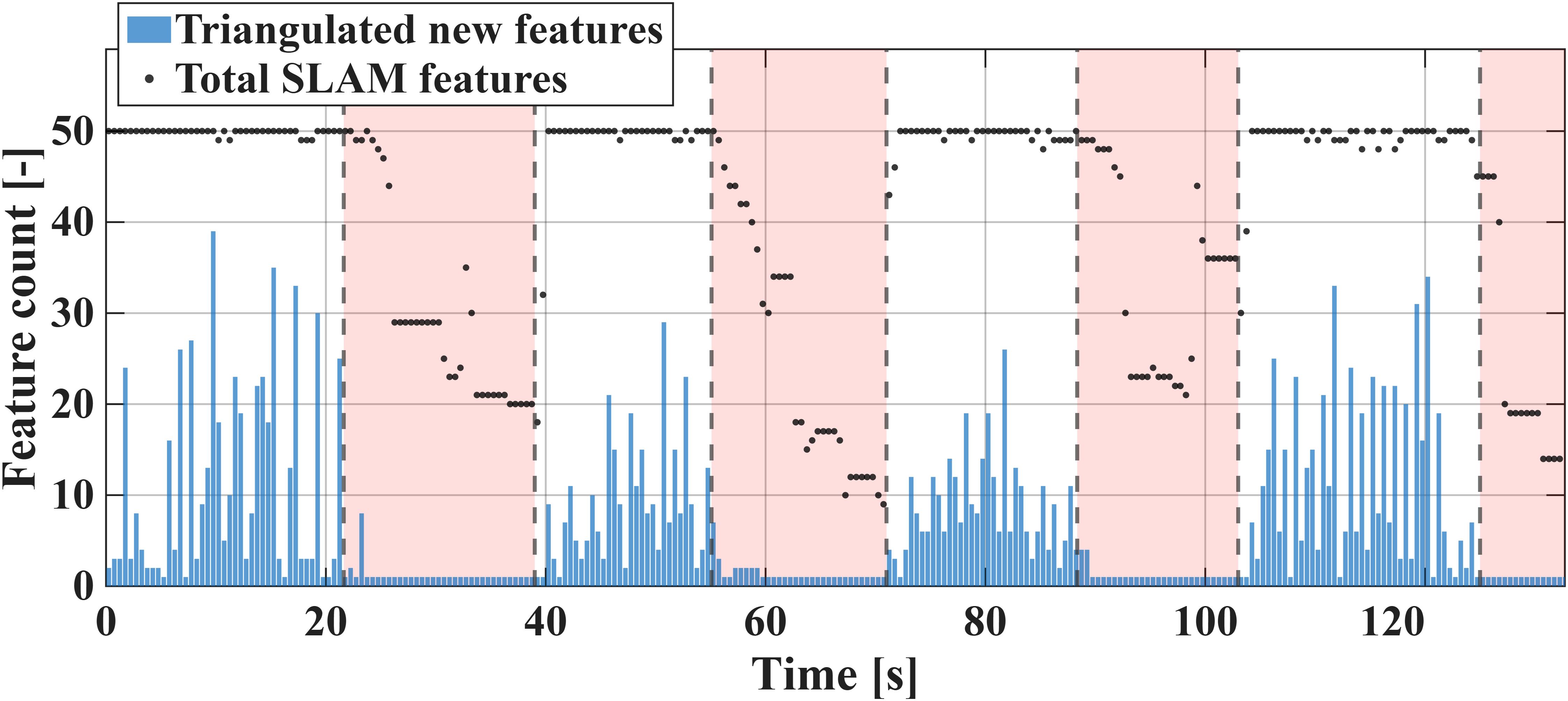}}
    \hfill
    \subfloat[Hard-CBF.\label{fig:square_hard_featureCount}]{%
        \includegraphics[width=0.32\textwidth]{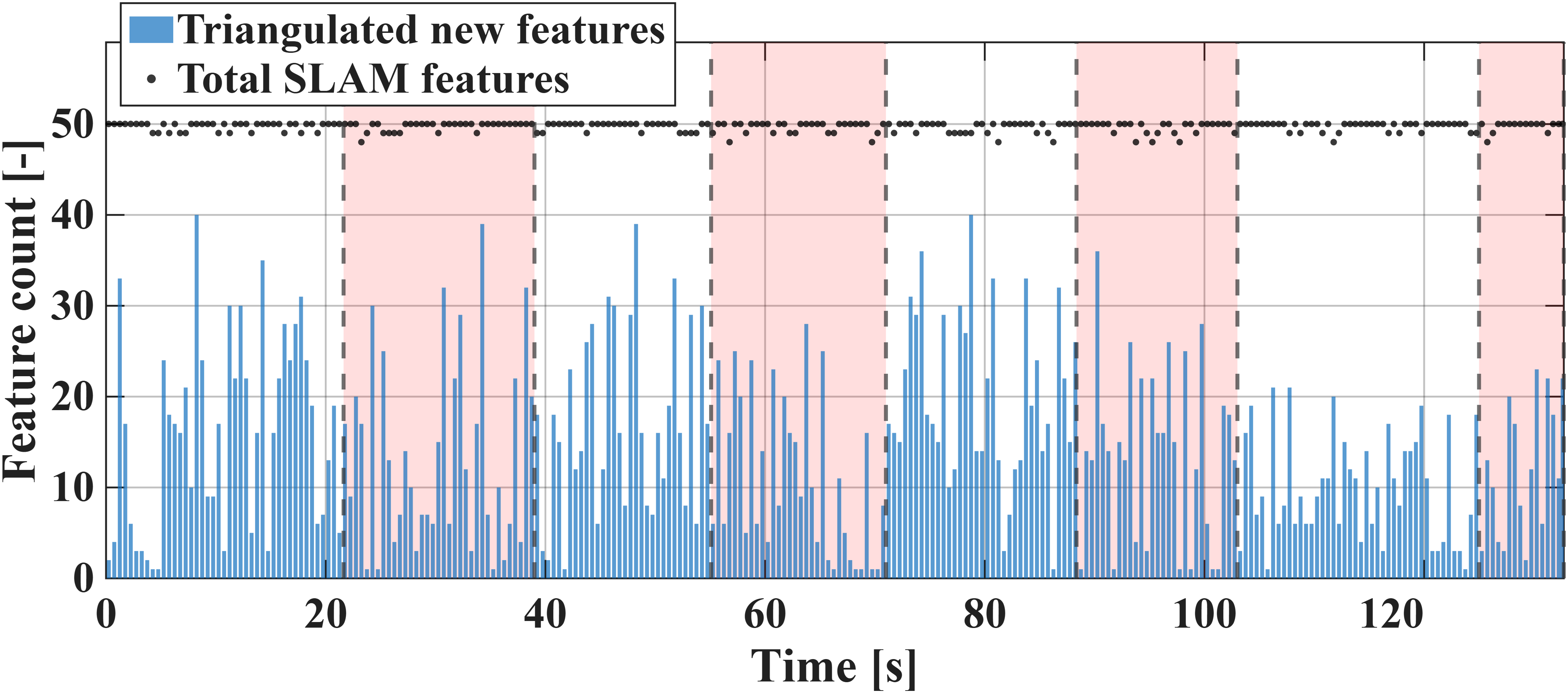}}
    \hfill
    \subfloat[Soft-CBF.\label{fig:square_soft_featureCount}]{%
        \includegraphics[width=0.32\textwidth]{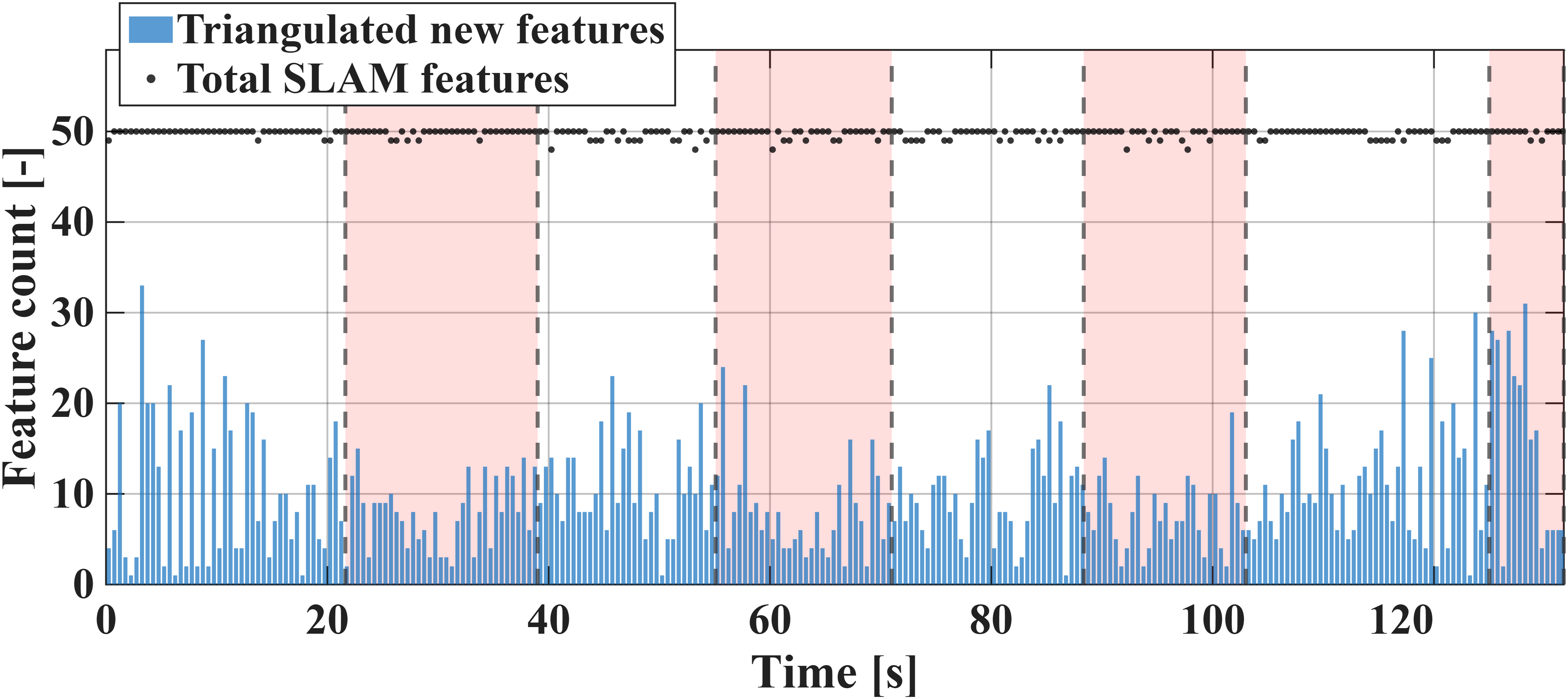}}
    \caption{Feature availability for the square trajectory under the nominal, hard-CBF, and soft-CBF cases. The shaded intervals denote the low-parallax portions identified from the nominal baseline and are overlaid on all cases as a common reference. The hard-CBF case increases newly triangulated features over these nominally degraded intervals, while the soft-CBF case preserves this benefit with reduced trajectory distortion.}
    \label{fig:square_feature_count_comparison}
\end{figure*}

\emph{Square trajectory:}
Fig.~\ref{fig:square_feature_count_comparison} compares the feature availability obtained under the nominal, hard-CBF, and soft-CBF cases for the square trajectory. In the nominal baseline, the number of newly triangulated features drops sharply in the shaded low-parallax intervals even though the total persistent-feature count can remain temporarily high. With the hard CBF, the additional motion commanded by the safety filter increases the number of newly triangulated features in these intervals. The soft-CBF case provides a less aggressive compromise: it maintains a usable supply of new triangulations while preserving the nominal path more closely than the hard constraint.

\begin{figure*}[!tb]
    \centering
    \subfloat[Nominal.\label{fig:climb_disabled_featureCount}]{%
        \includegraphics[width=0.32\textwidth]{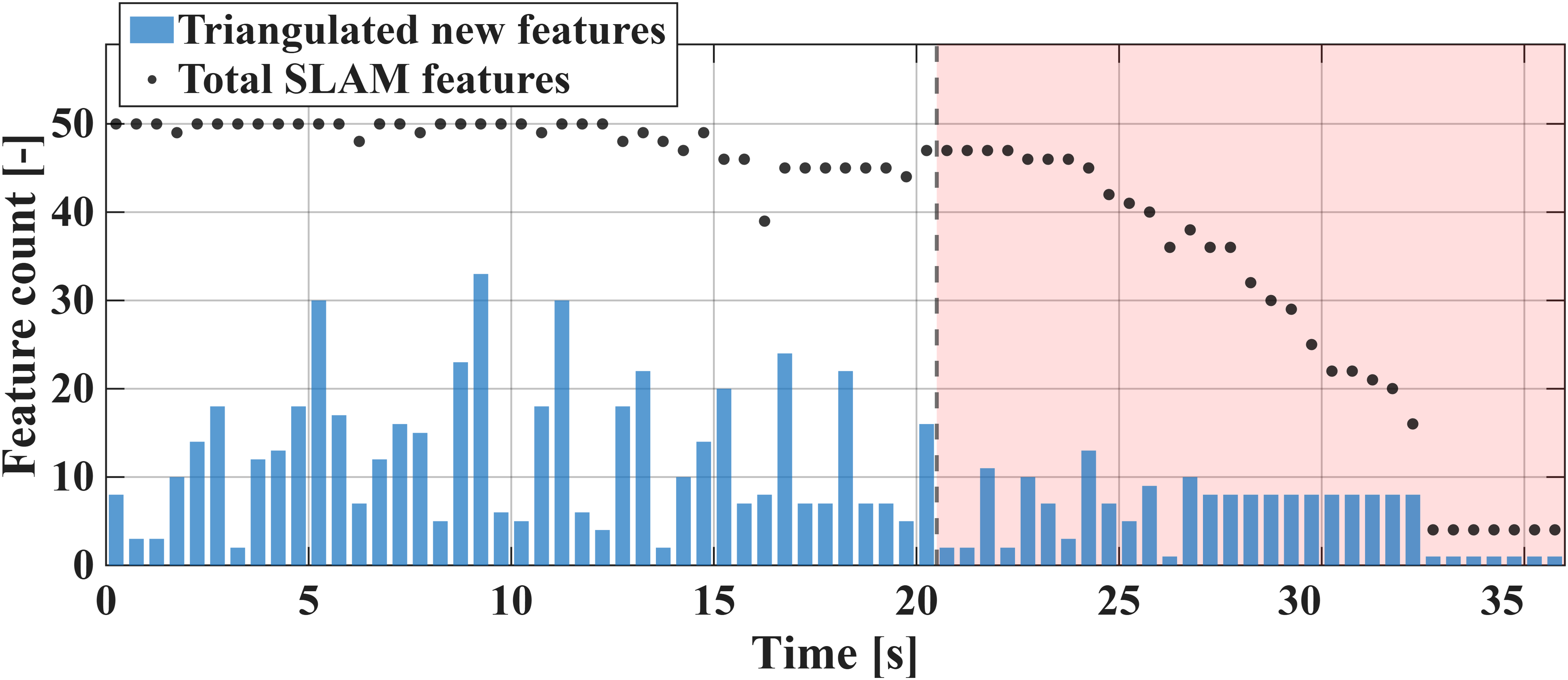}}
    \hfill
    \subfloat[Hard-CBF.\label{fig:climb_hard_featureCount}]{%
        \includegraphics[width=0.32\textwidth]{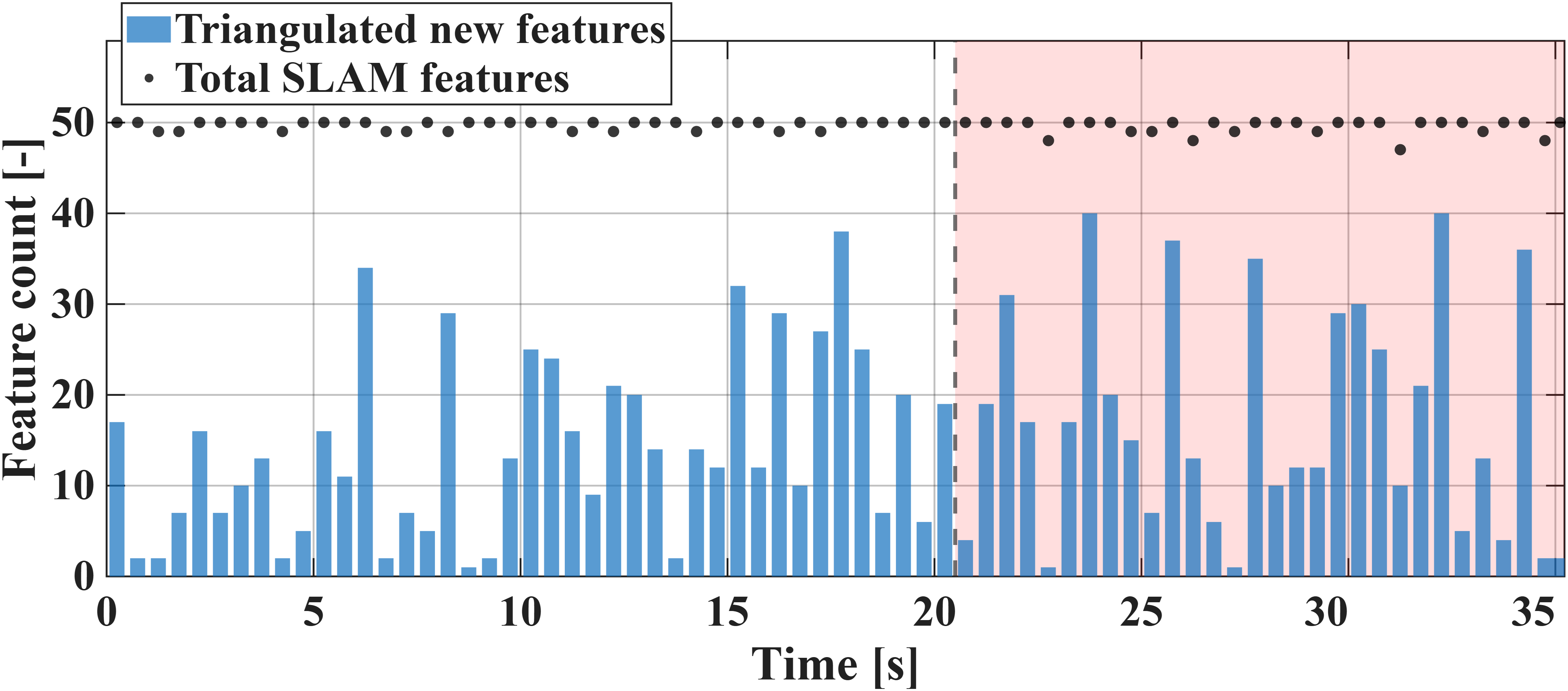}}
    \hfill
    \subfloat[Soft-CBF.\label{fig:climb_soft_featureCount}]{%
        \includegraphics[width=0.32\textwidth]{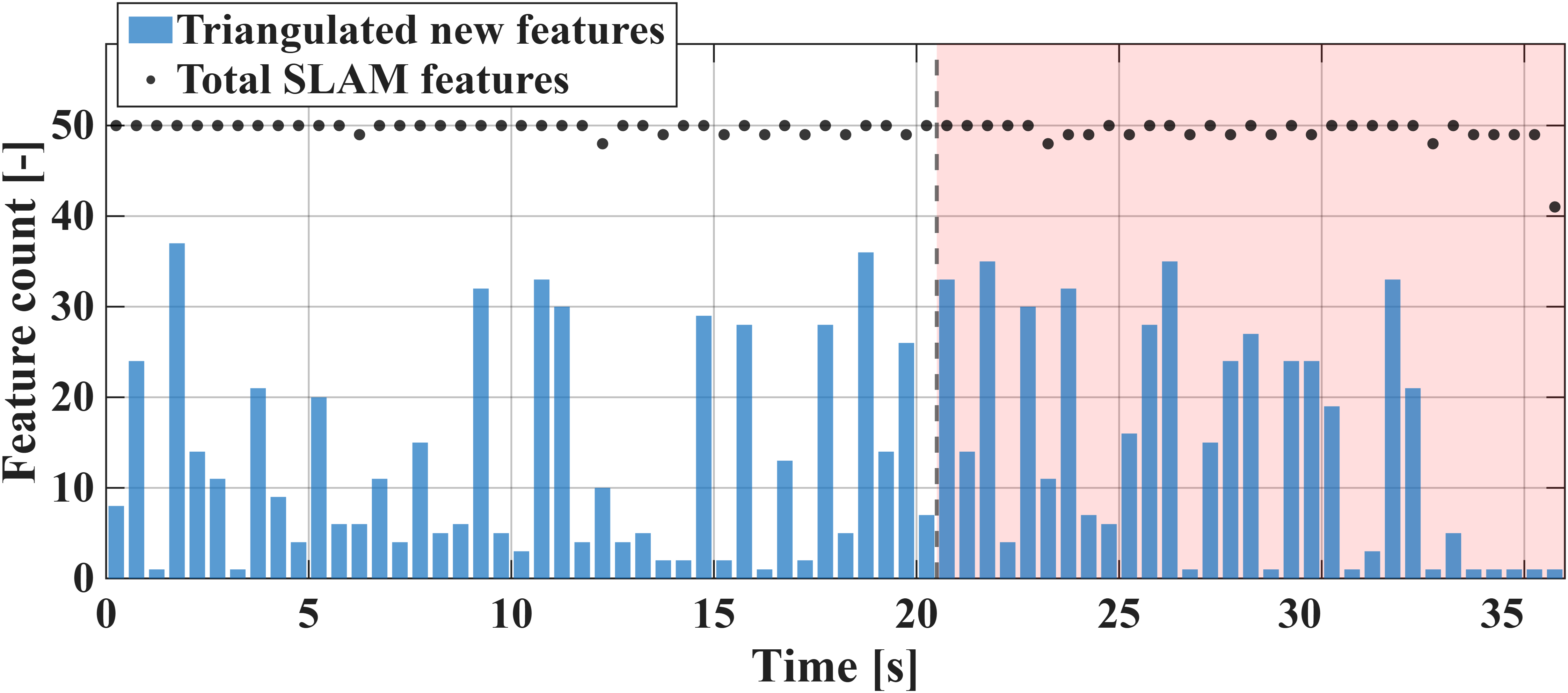}}
    \caption{Feature availability for the climbing trajectory under the nominal, hard-CBF, and soft-CBF cases. The shaded interval denotes the low-parallax portion identified from the nominal climb and is overlaid on all cases as a common reference to assess the effect of the CBF-induced motion. Both CBF configurations increase new feature triangulation over this nominally degraded interval. In the soft-CBF case, the terminal decrease is associated with the allowed near-hover or stop condition, for which continued parallax generation is no longer enforced as a strict requirement.}
    \label{fig:climb_feature_count_comparison}
\end{figure*}

\emph{Climbing trajectory:}
Fig.~\ref{fig:climb_feature_count_comparison} shows the corresponding feature statistics for the climbing trajectory. The nominal climb loses feature-observability. This degradation should not be attributed only to the terminal low-velocity phase; the increasing altitude also increases the translational motion required to preserve a sufficiently informative baseline. Both CBF configurations improve the supply of newly triangulated features by injecting additional motion when the nominal command becomes inadequate. The hard-CBF case provides the strongest triangulation-priority response, while the soft-CBF case retains the same qualitative benefit with reduced deviation from the nominal climb. The terminal reduction observed in the soft-CBF feature statistics is consistent with the permitted slowing or stopping behavior: once the vehicle approaches a near-hover state, little additional baseline is generated, and newly triangulated features may decrease even though the trajectory remains closer to the prescribed terminal condition.

Overall, to quantify the estimation-level consequence of the CBF-induced motion, the VIO trajectory is compared against the ground truth (SITL GPS) trajectory for the square and climb trajectory scenarios. Tables~\ref{tab:vio_error_square} and~\ref{tab:vio_error_climb} report the total ground-truth distance traveled and two VIO position-error metrics: the drift rate, computed as the ratio of the absolute trajectory error (ATE) root-mean-square error (RMSE) to the total ground-truth distance traveled, and the relative pose error (RPE), expressed as an RMSE over segments of duration $\Delta t=1~\mathrm{s}$.

\begin{table}[!tb]
\renewcommand{\arraystretch}{1.2}
\caption{VIO Position-Error Comparison for Square Trajectory on SITL.}
\label{tab:vio_error_square}
\centering
\begin{tabular}{l c c c}
\hline\hline
\textbf{Metric} & \textbf{Nominal} & \textbf{Hard-CBF} & \textbf{Soft-CBF} \\
\hline
GT distance (m)            & 401.7  & 510.5  & 425.3  \\
Drift (\%)                 & 2.610  & 0.227  & 0.324  \\
RPE RMSE (m)               & 0.528  & 0.136  & 0.198  \\
\hline\hline
\end{tabular}
\end{table}

\begin{table}[!tb]
\renewcommand{\arraystretch}{1.2}
\caption{VIO Position-Error Comparison for Climbing Trajectory on SITL.}
\label{tab:vio_error_climb}
\centering
\begin{tabular}{l c c c}
\hline\hline
\textbf{Metric} & \textbf{Nominal} & \textbf{Hard-CBF} & \textbf{Soft-CBF} \\
\hline
GT distance (m)            & 188.7  & 284.6  & 224.6   \\
Drift (\%)                 & 53.88  & 0.282  & 0.335  \\
RPE RMSE (m)               & 6.090  & 0.291  & 0.324  \\
\hline\hline
\end{tabular}
\end{table}

For the square trajectory, the number of triangulated features decreases during the turns but remains sufficient to prevent severe VIO divergence. Nevertheless, both CBF formulations further reduce the nominal drift. The improvement is most pronounced for the climbing trajectory, where the nominal case loses all triangulated features and consequently exhibits substantial drift. By preserving feature-observability, the proposed method maintains the drift at very low levels in both the hard- and soft-CBF cases. Consequently, the real-flight validation was focused exclusively on the climbing trajectory, as this scenario produced the most pronounced distinction between the nominal and CBF-assisted cases and therefore provided the most informative setting for experimentally evaluating the proposed method.

\begin{figure*}[!tb]
    \centering
    \subfloat[Multicopter platform.\label{fig:drone_body_view}]{%
        \includegraphics[width=0.48\textwidth]{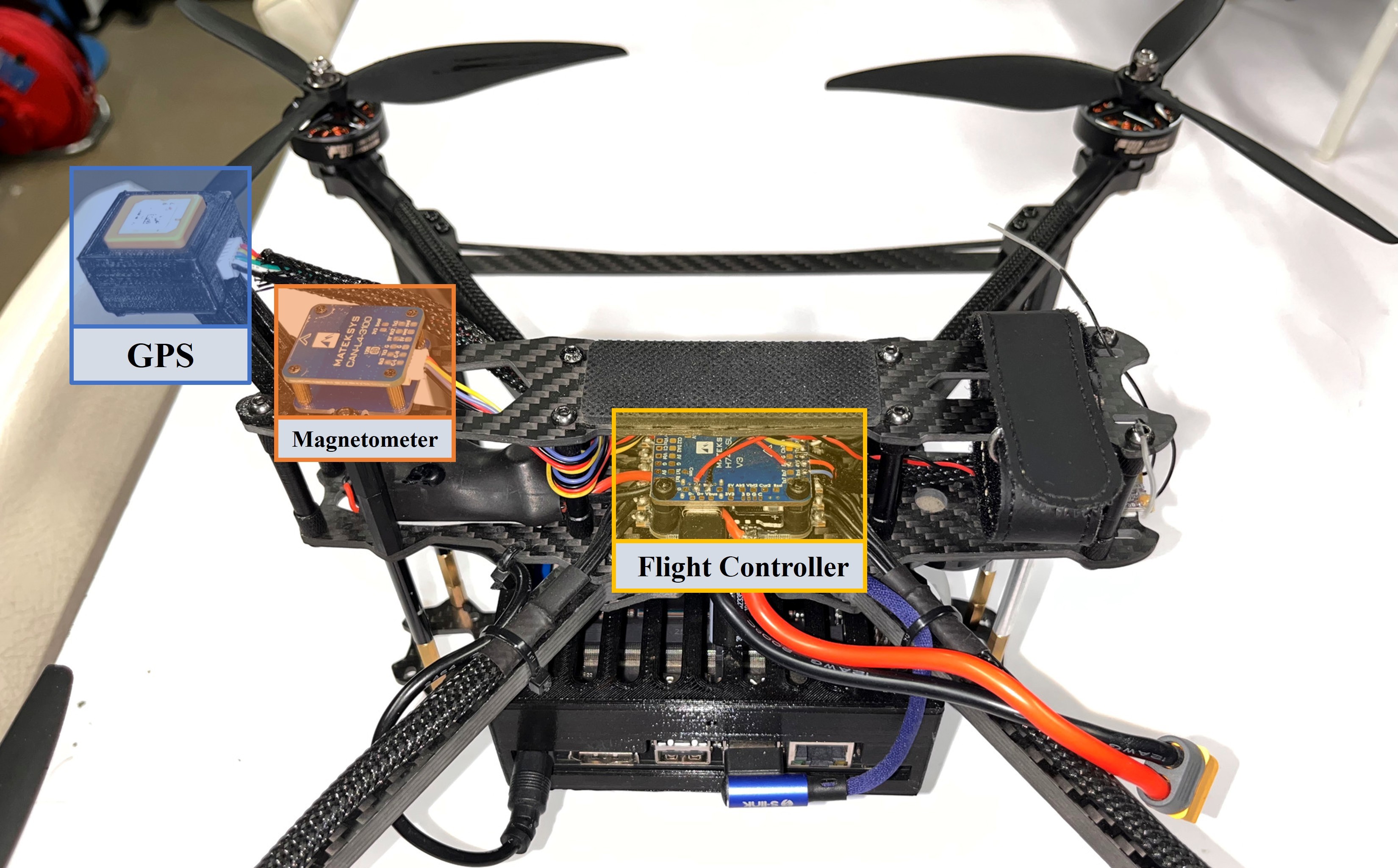}}
    \hfill
    \subfloat[Camera installation.\label{fig:drone_camera_view}]{%
        \includegraphics[width=0.48\textwidth]{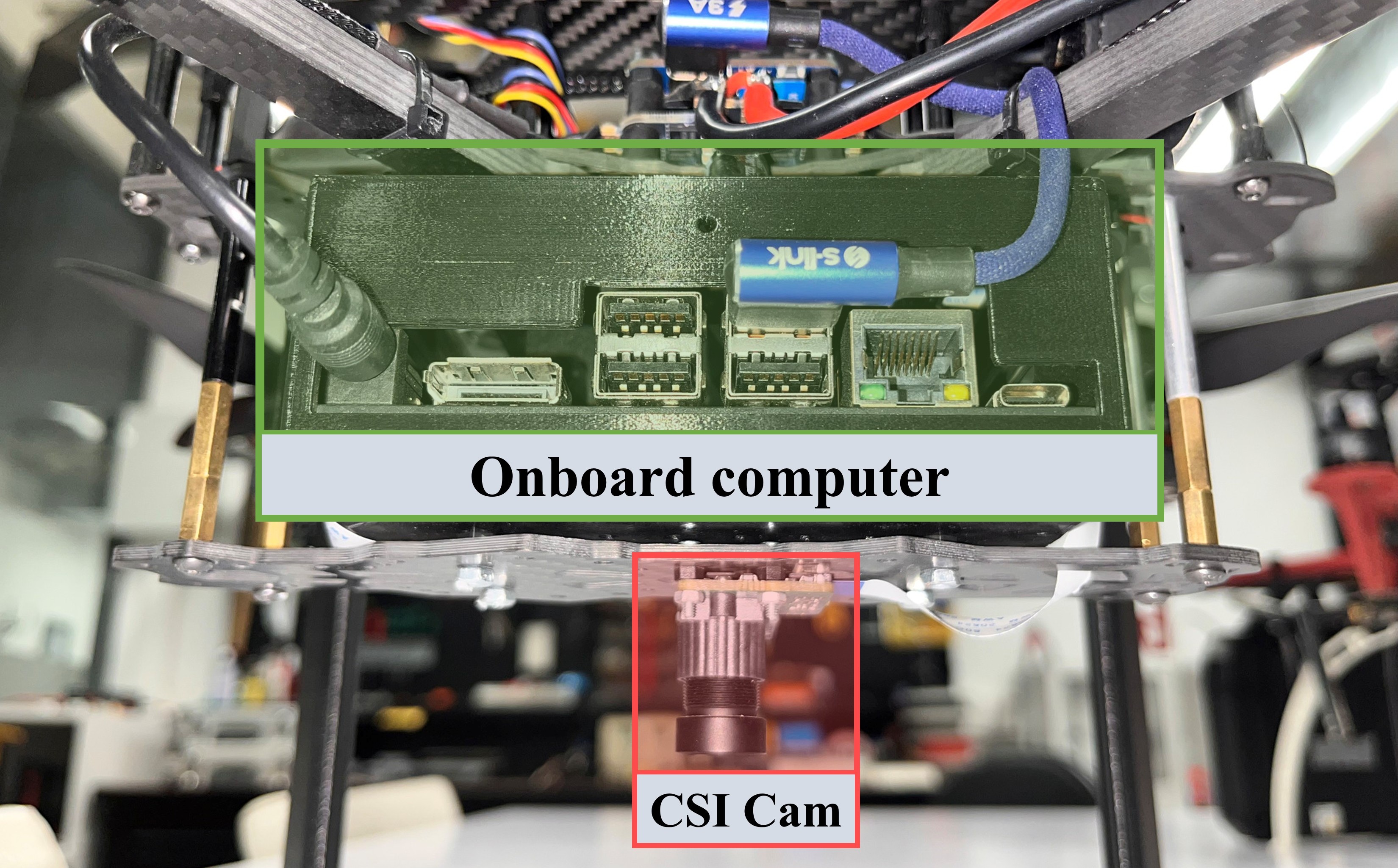}}
    \caption{Flight-test platform used for real-time evaluation of TANGO-VIO, including the onboard processing and downward-facing imaging hardware.}
    \label{fig:drone_view_group}
\end{figure*}

\subsection{Flight-Test Results}
\label{flightTests}

The flight experiments used the custom multicopter shown in Fig.~\ref{fig:drone_view_group}, equipped with a T-Motor propulsion system, MATEKSYS flight controller and magnetometer, NVIDIA Jetson Orin Nano, and a downward-facing $80^\circ$ field-of-view CSI camera. Only the climbing scenario was conducted in nominal and hard-CBF modes. Without retuning from SITL, TANGO-VIO was run at $20~\mathrm{Hz}$ with $\gamma=3$, $w_{\parallel}=1$, $w_{\perp}=10$, $\bar{\ell}_{\min}=-2.5$, and a $10~\mathrm{m/s}$ velocity-command limit.

\begin{figure*}[!tb]
    \centering
    \subfloat[Velocity command modification.\label{fig:test_climb_hard_velocityComparison}]{%
        \includegraphics[width=0.32\textwidth]{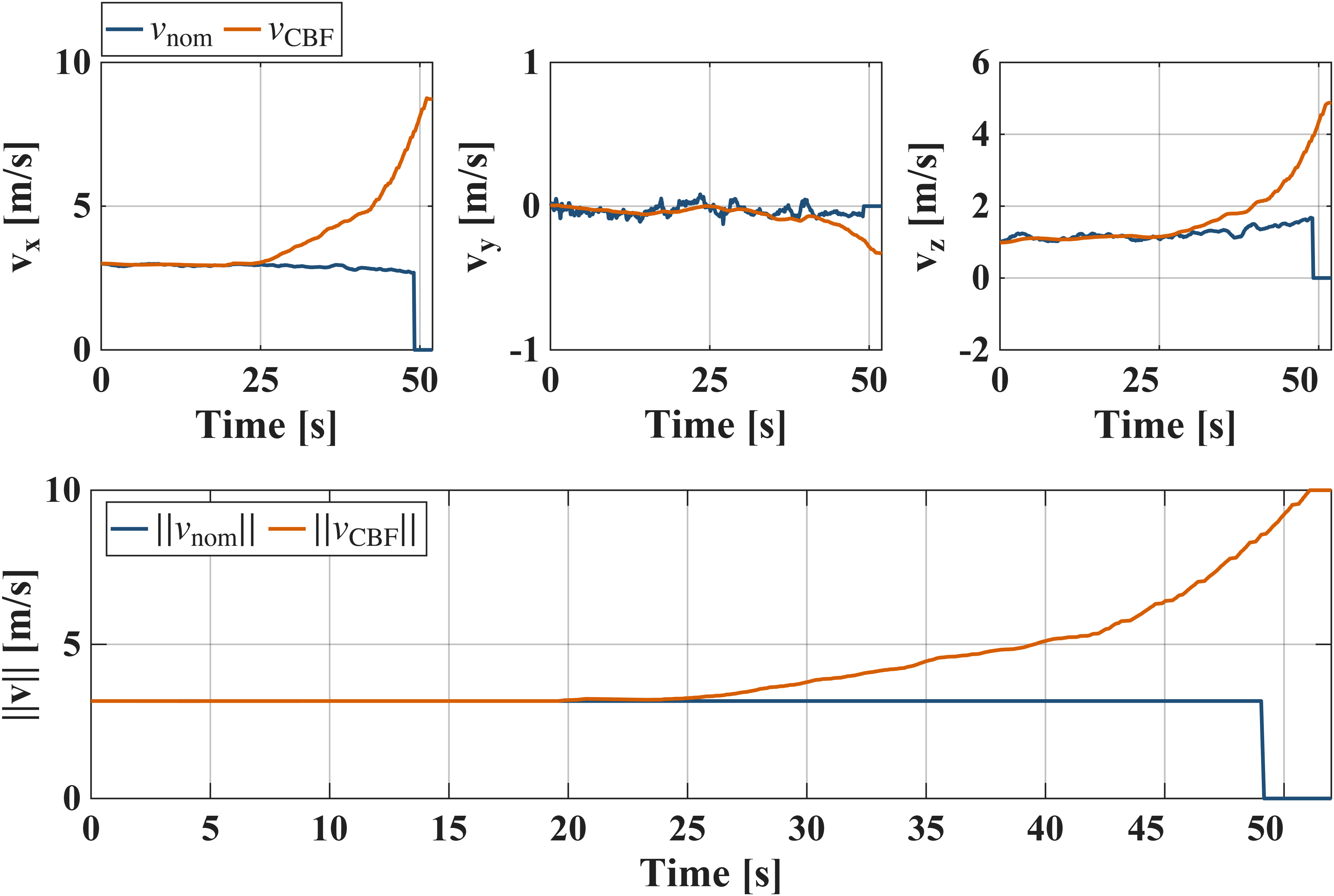}}
    \hfill
    \subfloat[Barrier-constraint history.\label{fig:test_climb_hard_CBFHistory}]{%
        \includegraphics[width=0.32\textwidth]{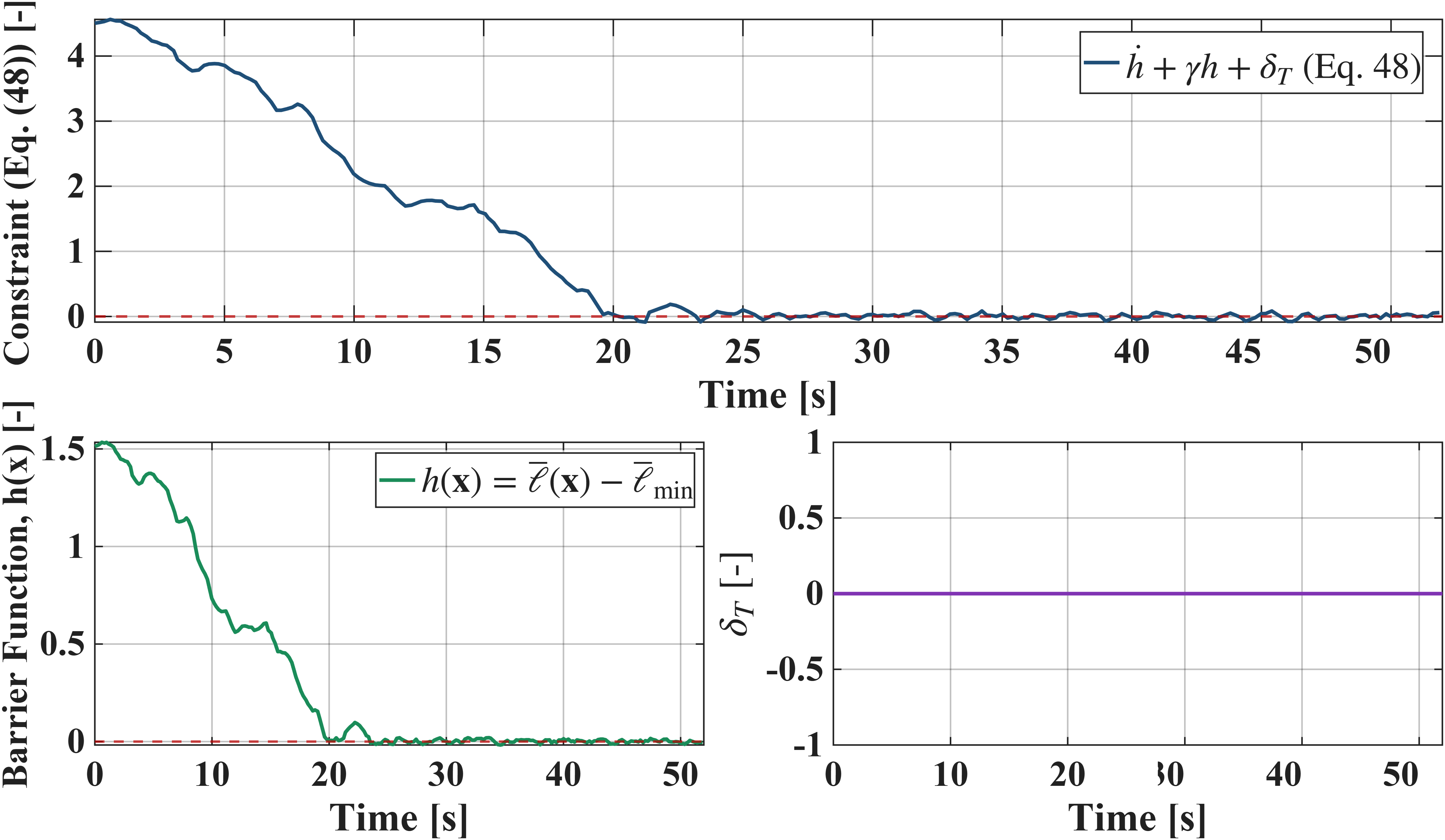}}
    \hfill
    \subfloat[Three-dimensional response.\label{fig:test_climb_hard_3Dtraj}]{%
        \includegraphics[width=0.32\textwidth]{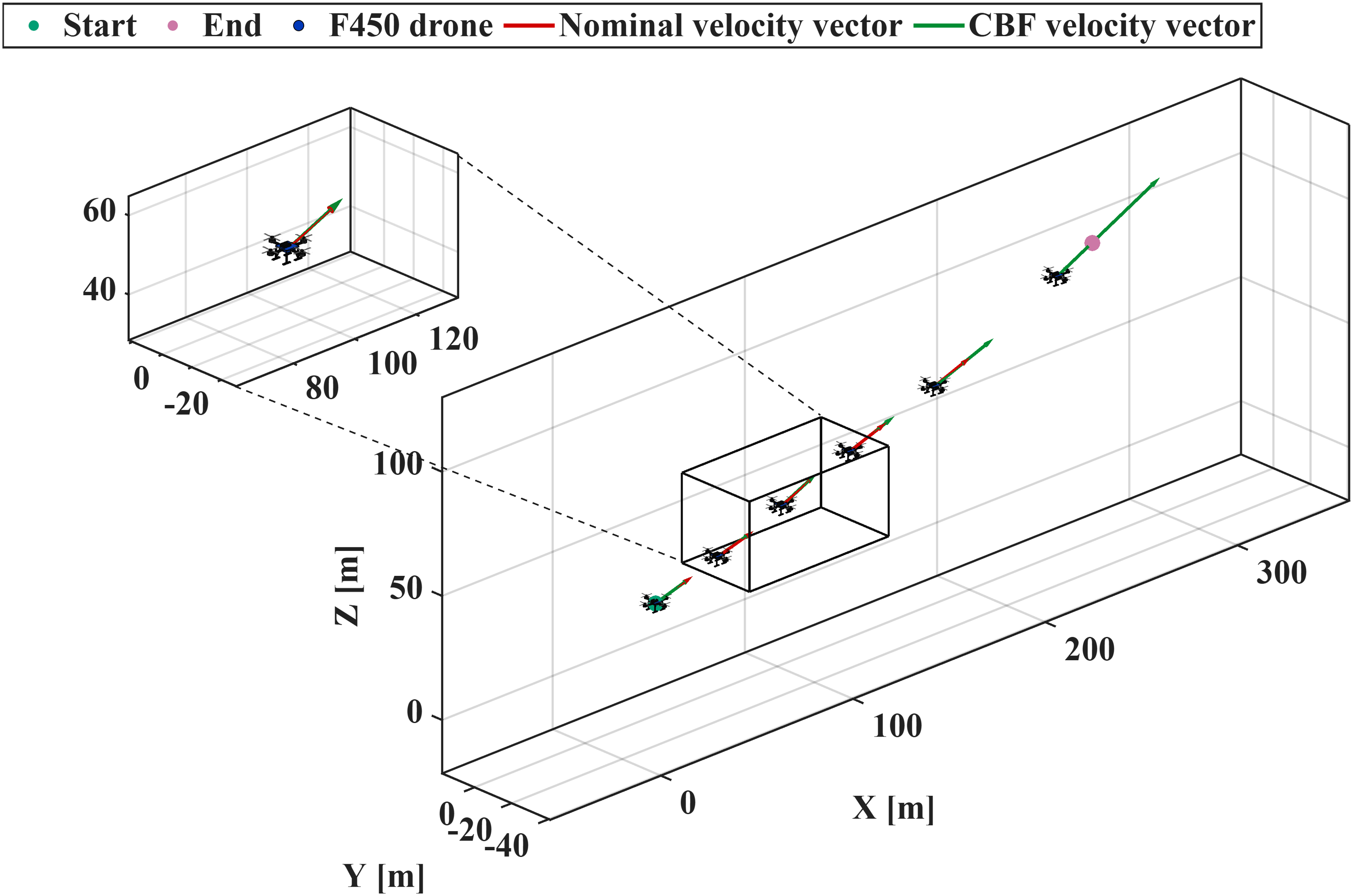}}
    \caption{Hard-CBF response in the climbing flight test. The additional motion maintains the triangulation-quality condition as altitude increases, with the associated trajectory deviation closely reproducing the SITL behavior.}
    \label{fig:test_climb_hard_cbf_response}
\end{figure*}

Fig.~\ref{fig:test_climb_hard_cbf_response} shows that the command-level response closely reproduces the corresponding SITL behavior. During the initial portion of the climb, the available parallax is sufficient and the corrected command remains close to the nominal command. As altitude increases, the same translational motion produces progressively weaker bearing variation, the barrier approaches its boundary, and the hard CBF becomes increasingly active. The resulting correction is concentrated primarily in the forward and vertical velocity components, while the lateral modification remains comparatively small. The corrected speed consequently increases toward the $10~\mathrm{m/s}$ limit, and the barrier and constraint residual are regulated near zero once the constraint becomes active. The accumulated correction produces the expected path and terminal-position deviation, consistent with the triangulation-priority response observed in SITL.

\begin{figure*}[!tb]
    \centering
    \subfloat[Nominal.\label{fig:test_climb_disabled_featureCount}]{%
        \includegraphics[width=0.48\textwidth]{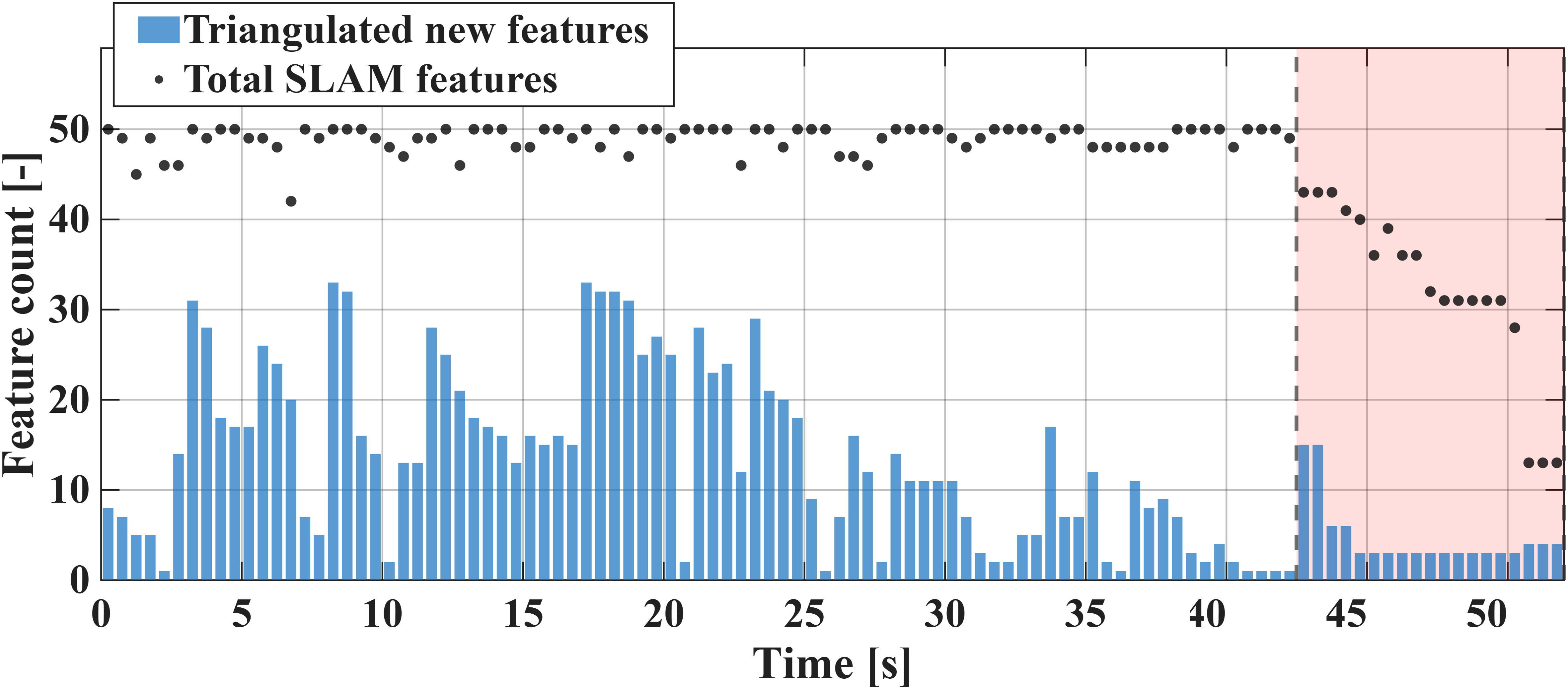}}
    \hfill
    \subfloat[Hard-CBF.\label{fig:test_climb_hard_featureCount}]{%
        \includegraphics[width=0.47\textwidth]{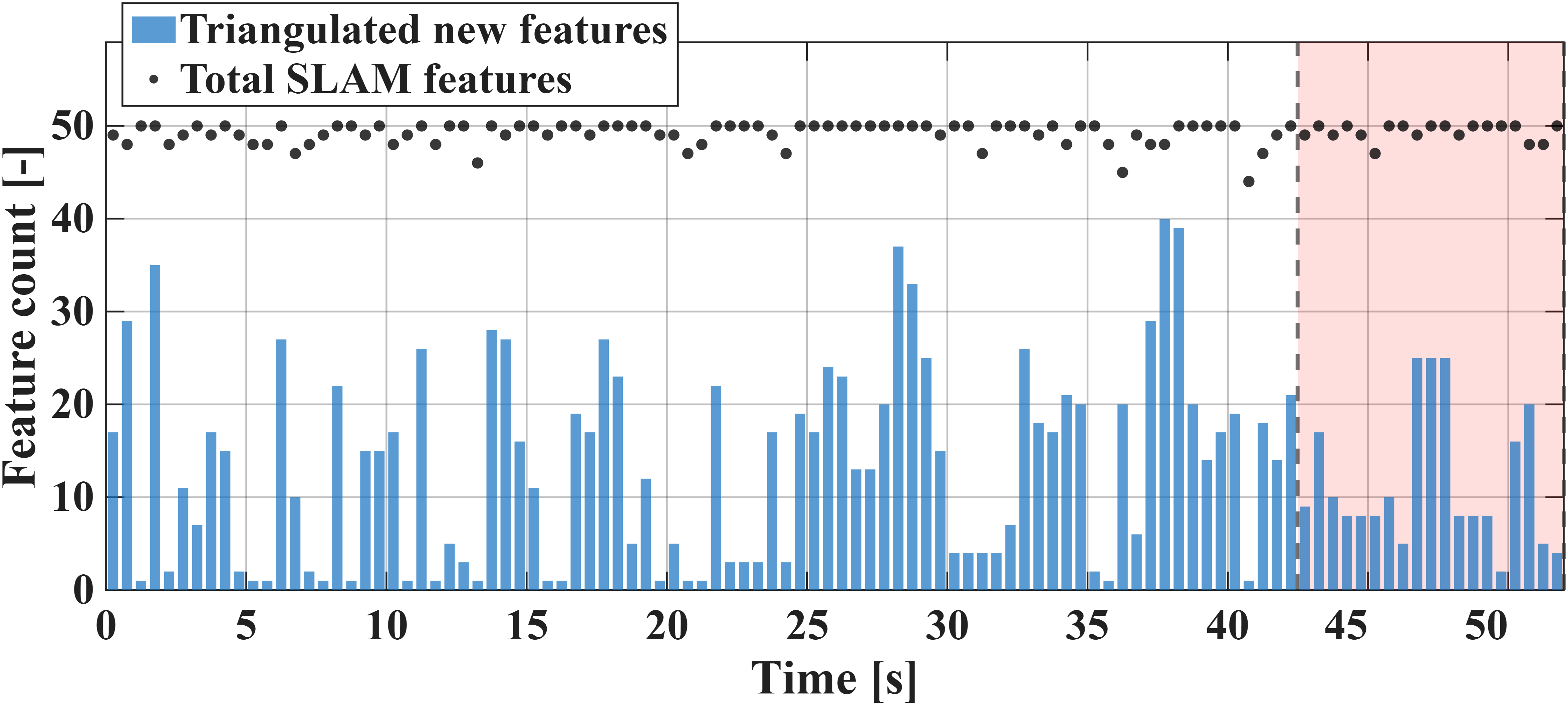}}
    \caption{Feature availability in the nominal and hard-CBF climbing flight tests. Over the common low-parallax interval, the hard-CBF motion sustains newly triangulated features and preserves the persistent-feature population, matching the trend observed in SITL.}
    \label{fig:test_climb_feature_count_comparison}
\end{figure*}

The corresponding feature statistics in Fig.~\ref{fig:test_climb_feature_count_comparison} confirm the visual-geometric consequence of this command modification. In the nominal flight, newly triangulated features become sparse during the highlighted low-parallax interval, followed by a pronounced reduction in the persistent (SLAM) feature population. Under the hard CBF, the additional parallax-generating motion sustains new feature initialization and keeps the persistent-feature population close to its available maximum over the same interval. Because the flight implementation uses the SITL parameterization without retuning, this agreement shows that the mechanism identified in simulation is retained under real sensing, estimation, feature tracking, and vehicle-response effects. The flight tests therefore reproduce both aspects of the SITL result: improved triangulation quality through additional motion and the associated loss of trajectory fidelity under strict enforcement.

\section{Conclusion}
\label{conclusion}

Insufficient parallax remains a critical limitation for visual-inertial odometry because low-speed, near-static, or geometrically uninformative motion can degrade bearing-based feature triangulation and the feature-observability required for reliable operation. This study addressed this issue through TANGO-VIO, a triangulation-aware navigation framework that actively regulates the underlying feature geometry. A feature-wise stacked-bearing matrix was used to define an aggregate log-determinant triangulation-quality metric, which was embedded in a CBF and enforced directly at the body-frame velocity-command level. The resulting safety filter preserves the nominal command while the prescribed condition is satisfied and introduces corrective, parallax-generating motion when required. Hard- and soft-CBF formulations were developed to represent strict threshold enforcement and controlled relaxation when triangulation-quality maintenance conflicts with the nominal mission objective.

The SITL results showed that the resulting velocity corrections arise consistently from the geometry of the executed motion rather than from arbitrary trajectory deformation. In the square scenario, degradation was concentrated around the low-velocity turns, where the available translational baseline was insufficient for continued feature triangulation. In the climbing scenario, increasing altitude progressively increased the translation required to generate an informative bearing change, causing the constant nominal command to become inadequate. In both cases, the hard CBF maintained the prescribed barrier condition at the cost of larger path and terminal deviations. The soft CBF retained the same triangulation-aware behavior while remaining closer to the nominal trajectory through explicitly quantified relaxation. This distinction was particularly evident near terminal stop or hover conditions, which constitute a general conflict between trajectory completion and continued parallax generation. The feature statistics supported the command- and trajectory-level results: the CBF-induced motion increased newly triangulated features and preserved the persistent feature population over intervals in which the nominal motion became visually uninformative. The climbing flight tests closely reproduced this cause--effect behavior using the same parameterization. The nominal flight exhibited declining new triangulations and persistent-feature availability in the low-parallax interval, whereas the hard CBF sustained both through additional motion and produced the trajectory deviation predicted by the SITL analysis.

Consequently, this work formulates feature-observability-aware motion generation as a real-time safety-filtering problem and establishes a direct link between bearing geometry, stacked-bearing matrix conditioning, triangulation-quality preservation, and CBF-certified velocity correction. The hard-CBF mode is appropriate when maintaining the prescribed feature-observability condition has priority over trajectory fidelity, whereas the soft-CBF mode is better suited to missions requiring a balance among visual-information quality, path tracking, and terminal behavior. The close SITL--flight agreement demonstrates the practical realizability of the proposed mechanism and establishes the hard-constrained TANGO-VIO implementation as flight-proven for the tested climbing scenario. Overall, TANGO-VIO converts feature-geometric degradation from a passive consequence of the commanded trajectory into an actively regulated navigation constraint.

Future work will investigate an adaptive log-determinant threshold to reduce path distortion while preserving a minimum triangulation-quality floor, and an uncertainty-aware metric that incorporates feature-tracking quality and measurement reliability.

\ifCLASSOPTIONcaptionsoff
  \newpage
\fi

\printbibliography[title=References,category=cited]

\end{document}